\documentclass[11pt]{article}
\usepackage[top=2cm, bottom=2cm, left=2cm, right=2cm, heightrounded,
  marginparwidth=2.9cm, marginparsep=2mm]{geometry}

\usepackage{amsfonts,mathrsfs}
\usepackage{textcomp}
\usepackage{xcolor}
\usepackage{xspace}
\usepackage{hyperref}
\usepackage[ruled,vlined,noend]{algorithm2e}
\usepackage{caption}
\usepackage{subcaption}
\usepackage{graphicx}
\usepackage{lipsum}
\usepackage{multicol,multirow}
\usepackage{booktabs}
\usepackage{amsmath, nccmath}
\usepackage{bbm}
\usepackage{enumitem}

\usepackage[compact]{titlesec}
\titlespacing{\section}{0pt}{2ex}{1ex}
\titlespacing{\subsection}{0pt}{1ex}{0ex}
\titlespacing{\subsubsection}{0pt}{1ex}{0ex}

\DeclareMathOperator*{\argmax}{arg\,max}
\DeclareMathOperator*{\argmin}{arg\,min}

\makeatletter
\DeclareRobustCommand\onedot{\futurelet\@let@token\@onedot}
\def\@onedot{\ifx\@let@token.\else.\null\fi\xspace}

\def\eg{\emph{e.g}\onedot} 
\def\ie{\emph{i.e}\onedot} 
\def\cf{\emph{cf}\onedot}

\def\ow{{\rm else}}

\makeatother

\title{A BIC-based Mixture Model Defense against\\
        Data Poisoning Attacks on Classifiers}

\author{Xi Li, David J. Miller, Zhen Xiang and George Kesidis\\
School of EECS, Pennsylvania State University, University Park, PA, 16802, USA\\
\{xzl45, djm25, zux49, gik2\}@psu.edu}

\begin{document}

\maketitle

\begin{abstract}
Data Poisoning (DP) is an effective attack that causes trained classifiers to misclassify their inputs. DP attacks significantly degrade a classifier's accuracy by covertly injecting attack samples into the training set.
Broadly applicable to different classifier structures, 
without strong assumptions about the attacker,
an {\it unsupervised} Bayesian Information Criterion (BIC)-based mixture model defense against
``error generic" DP attacks 
is herein proposed that: 
1) addresses the most challenging {\it embedded} DP scenario wherein, if DP is present, the poisoned samples are an {\it a priori} unknown subset of the training set, and with no clean validation set available;
2) applies a mixture model both to well-fit potentially multi-modal class distributions and to capture poisoned samples within a small subset of the mixture components; 3) jointly identifies poisoned components and samples by minimizing the BIC cost defined over the whole training set, with the identified poisoned data removed prior to classifier training. 
Our experimental results, for various classifier structures and benchmark datasets, demonstrate the effectiveness and universality of our defense under strong DP attacks, as well as its superiority over other works.
\end{abstract}

\section{Introduction}

Learning-based models have shown impressive performance in various domains, \eg, computer vision and natural language processing. However, machine learning systems are vulnerable to maliciously crafted inputs. Interest in Adversarial Learning (AL) has grown dramatically in recent years, focused on devising attacks against machine learning models and defenses against same. Three important types of AL attacks are \cite{Review}: data poisoning  (\eg, \cite{DBLP:journals/corr/abs-1712-05526, DBLP:journals/ijon/XiaoBNXER15, DBLP:conf/ijcai/Ma0H19, DBLP:conf/nips/LiWSV16, DBLP:conf/aaai/AlfeldZB16, BadNets2019, Trojan}), test-time evasion (\eg, \cite{Biggio_2013,DBLP:conf/eurosp/PapernotMJFCS16,attack-graph}), and reverse engineering (\eg, \cite{DBLP:conf/iclr/OhAFS18, DBLP:conf/ccs/PapernotMGJCS17}). In this work, we address Data Poisoning (DP) attacks against models trained for classification tasks.

DP attacks are training-time attacks which
%(\eg, \cite{DBLP:journals/corr/abs-1712-05526, BadNets2019, DBLP:journals/ijon/XiaoBNXER15, ZhenBackdoor2019, DBLP:conf/ijcai/Ma0H19, DBLP:conf/nips/LiWSV16, DBLP:conf/aaai/AlfeldZB16}) 
involve the insertion of ``poisoned" samples into the training set of a classifier. 
In this paper, we address the so-called ``error generic" DP attacks (hereafter called DP attacks) \cite{Biggio_2018} which aim to degrade the {\em overall} classification accuracy\footnote{Error specific attacks, particularly backdoor attacks involving specific backdoor patterns and source and target classes, \eg, \cite{DBLP:journals/corr/abs-1712-05526, Trojan, BadNets2019}, are not the focus herein.}.
To effectively mislead classifier training using relatively few poisoned samples, an attacker introduces feature collision\cite{Hong20GS} to the training samples by \eg flipping the class labels of clean samples. The information extracted from the clean and poisoned samples labeled to the same class contradicts each other and prevents the learning of an accurate class decision boundary.
DP attacks have been successfully demonstrated against 
Support Vector Machines (SVMs) \cite{DBLP:journals/ijon/XiaoBNXER15}, 
Logistic Regression (LR) models \cite{DBLP:conf/nips/FengXMY14}, auto-regressive models \cite{DBLP:conf/aaai/AlfeldZB16}, collaborative filtering systems \cite{DBLP:conf/nips/LiWSV16}, differentially-private learners \cite{DBLP:conf/ijcai/Ma0H19}, and neural networks (NN) \cite{Munoz-GonzalezB17}. 
% Generally, the defense proposed here can be deployed to sanitize the training set prior to the training process.

In the most general setting (and also the most challenging one for the defender), considered here, the poisoned samples, {\it if there is data poisoning}, are an {\em unknown} subset embedded among the clean training samples. 
% We propose a BIC-based mixture model defense strategy against error-generic DP attacks, where poisoned samples are {\em unknown} subset embedded among clean data samples. 
That is, the defender does not know whether an attack is present, and if so, which samples are poisoned and which class(es) are corrupted. This {\it embedded data poisoning} attack scenario is of great practical interest, and yet remains largely unsolved. Studies on defending against such attacks either are tailored to a specific type of classifier (\eg, SVM \cite{curie}, LR \cite{DBLP:conf/nips/FengXMY14}), are only suitable for a specific type of DP attack (\eg, \cite{PaudiceML18} only defends against label flipping attacks), or make strong assumptions about the training data (\eg, availability of a clean validation set for use by the defender 
\cite{RONI}).
%\cite{RONI,mixture}).   
The proposed method does not make any such assumptions about the attack,
does not require a clean (attack-free) validation set, and can be deployed to protect various types of classifiers.

Poisoned samples are generally {\em atypical} of the distribution of the class to which they are labeled. We thus apply mixture modeling \cite{Duda,Mclachlan} to accurately explain the potentially multi-modal data and to capture poisoned samples within a subset of mixture components. 
We make the following observations.
If the poisoned samples are typical of another class
(different from the class to which they are labeled),
we expect that re-distributing them to other classes should increase the overall data likelihood.
Furthermore, removing a poisoned component will reduce the model complexity of a mixture.  Thus, both the data likelihood and model complexity terms that constitute the Bayesian Information Criterion (BIC) objective function \cite{schwarz1978} should improve when data poisoning is mitigated.  
Accordingly, we propose to make poisoned sample inferences consistent with minimizing BIC. 
% To accurately explain the (possibly multi-modal) class data, 
We first apply mixture modelling separately to each class, with the number of components chosen to minimize the BIC criterion. 
% The optimal number of mixture components is selected by minimizing BIC cost. 
Then we assess components for possible poisoning, with a detected component either removed or revised (whichever results in a lower BIC cost). After poisoned samples have been detected and removed, the classifier is trained on the resulting (sanitized) data set.

In summary, our BIC-based mixture model defense is:
\begin{itemize}
    \item \textit{Novel}: As far as we know, we are the first to formulate a BIC-based defense for \textit{unsupervised} anomaly detection/DP attack mitigation.
    \item \textit{Practical}: We address the practical and challenging embedded DP attack scenario, and wherein more than one class may be poisoned. Moreover, our defender does not require a clean validation set.
    \item \textit{Robust}: The experimental results on several benchmark datasets demonstrate the effectiveness of our defense under strong DP attacks, as well as the superiority over other works. In principle, our method can be deployed to defend against any error-generic DP attack. 
    \item \textit{Comprehensive}: Our approach sanitizes the training set (prior to training) and is applicable to a wide variety of classifier structures, including SVMs, LR and NN-based classifiers, as will be seen in the sequel. 
\end{itemize}

The rest of the paper is organized as follows: We first review several existing DP studies in Section \ref{related work}. Then we define our threat model for DP attacks in Section \ref{threat model}. In Section \ref{method}, we propose our BIC-based mixture model defense. Experimental results on binary classification tasks and multi-class classification tasks are presented in Section \ref{experiments_bi} and Section \ref{experiments_multi}, respectively. Finally, we summarize our work in Section \ref{summary}.

\section{Related Work}\label{related work}

An obvious strategy for defending against data poisoning attacks is to conduct ``data sanitization'' on the training set, \ie, identifying and cleansing the attack samples as training set outliers.
Outlier identification is divided into two types -- supervised and unsupervised.
% \cite{DBLP:journals/air/HodgeA04, DBLP:journals/csur/ChandolaBK09}. 
Some supervised detection methods are in fact a form of supervised learning -- the defender trains a binary discriminator based on labeled examples of anomalies and normalities, \eg, \cite{DBLP:journals/access/YiZSD18}. The resulting discriminator can then be applied to identify and sanitize anomalies (possibly poisoned examples) in the classifier's training set.  However, such a learned discriminator may only reliably identify anomalies in a data set that are similar to those seen during the discriminator's training -- {\it i.e.}, the discriminator may only be good at identifying {\it known} anomalies,
not {\it unknown} anomalies.  Another issue with this approach is that
anomalies (attack instances) may be rarer and more difficult to collect than ``normalities''.  Thus, the performance of the discriminator may suffer from highly skewed class imbalance (``anomalous'' vs. ``normal'') in the supervised examples used to train it.

Other supervised detection methods are more akin to {\it unsupervised} anomaly detection methods, except that they possess hyper-parameters whose setting, to achieve good detection performance, requires either a clean validation set (which is anathema to the embedded DP scenario considered here) or, again, a labeled set of ``normalities'' and ``anomalies'', \eg, \cite{DBLP:conf/sigmod/SongGHW21, DBLP:conf/iclr/RaghunathanSL18,DBLP:conf/nips/SteinhardtKL17,PaudiceML18,DiakonikolasKK019,Hong20GS}.
On the other hand, truly unsupervised detection methods do not require labeled examples of what is normal and what is anomalous, and are analogous to unsupervised clustering methods. They model data distributions and flag potential outliers, \eg, \cite{DBLP:conf/sigmod/ChaiC00LM20, mixture, 9247440, 9507359} and our method proposed herein. The false positive rate (\ie, the fraction of normal samples misidentified as outliers) and false negative rate (\ie, the fraction of true outliers misidentified as normalities) may be relatively high for an unsupervised detector.  Thus, \cite{DBLP:conf/sigmod/ChaiC00LM20} proposed to leverage human intelligence to correctly identify outliers from amongst a set of outlier candidates detected by a machine-learning method.  However, such an approach is only suitable for domains where humans are skilled at analyzing data -- images, speech, text, low-dimensional data domains, and/or high-dimensional ones that can be visualized with fidelity in a low-dimensional space.  Moreover, such an approach may be very costly and time-consuming, as there may be a large set of outlier candidates for the human to inspect.

% For semi-supervised detection, the defender only requires that some normal samples are correctly labeled; that is, the outliers are not {\it a priori} labeled as anomalies. Thus, semi-supervised detection is more widely applicable, \eg, \cite{DBLP:conf/sigmod/SongGHW21, DBLP:conf/iclr/RaghunathanSL18,DBLP:conf/nips/SteinhardtKL17,PaudiceML18,DiakonikolasKK019,Hong20GS,mixture} and our method herein.

Due to the near-limitless space of possible attacks, in practice there will always be {\it unknown} attacks, ones which have not been encountered before (and thus, with no labeled examples available for training of a supervised detector).  For various types of adversarial attacks (test-time evasion attacks, as well as data poisoning attacks), it has been observed that the performance of supervised detectors may fare poorly on unknown attacks \cite{Review}.
On the theoretical front, there are works such as \cite{DBLP:conf/nips/SteinhardtKL17,DBLP:conf/iclr/RaghunathanSL18}.
Given a defender that first performs outlier removal followed by margin-based loss minimization,  \cite{DBLP:conf/nips/SteinhardtKL17} generates an approximate upper bound on the efficacy of any DP attack.
They also established a dual method which generates an attack that nearly achieves this upper bound.  However, their attack requires full knowledge of the clean training set (prior to its poisoning) and cannot handle non-convex loss functions, which limits its application in practice. \cite{DBLP:conf/iclr/RaghunathanSL18} proposes a method for producing certificates of robustness for two-layer neural networks. Such certificates are differentiable and can be jointly optimized with the network parameters, providing an adaptive regularizer that encourages robustness against attacks. However,
they neither characterize nor discuss the inherent tradeoff between robustness and model bias ({\it i.e.}, the degradation in the classifier's accuracy on {\it clean} data that results from making the classifier robust to attacks). 

% \cite{DBLP:conf/sigmod/SongGHW21,PaudiceML18} keep outliers in the dataset after data sanitization.
% \cite{DBLP:conf/sigmod/SongGHW21} detects outliers for clustering algorithms and saves identified outliers for better clustering by (minimally) adjusting their erroneous values.  
\cite{PaudiceML18}
proposed a label sanitization strategy tailored to label flipping attacks. The poisoned samples 
% (\ie, samples with flipped labels) 
are expected to be outliers relative to untainted samples with the same labels. Thus, they relabel a sample based on the plurality label of its $K$ nearest neighbors (KNN) to enforce label homogeneity. However, 
% it is effective only under weak attacks. 
this defense will fail when the number of poisoned samples is sufficiently large such that some of the neighbors of an attack sample are also attack samples. This defense also relies on the availability of a clean validation set to tune the hyper-parameter $K$. This choice highly impacts the detector's performance, as will be seen (\cf Section \ref{experiments_bi} and \ref{experiments_multi}).
% and the plurality threshold $\eta$. 

\cite{DiakonikolasKK019,Hong20GS}
proposed to defend DP attacks by analyzing training sample gradients (measured with respect to the loss function used for classifier training). They posited a unified view of effects of DP on learned classifier parameters: (1) the $l_2$ norm of the gradient from a poisoned sample is larger than that of a clean sample, on average; (2) there is an orientation difference between poisoned and clean sample gradients. 
\cite{DiakonikolasKK019} detects such effects by singular value decomposition (SVD) applied to the matrix of gradients, with each row of this matrix the gradient, with respect to the model parameters, of one sample's contribution to the training loss function.
They derive an outlier score for each training sample, which is the squared magnitude of the projection of the gradient onto the top right singular vector.
For classification tasks, \cite{DiakonikolasKK019} separately constructs the matrix of gradients for the training samples from each class, and computes outlier scores for each class, to improve performance.
At each detection step, the top $\frac{\epsilon}{\beta}$ fraction of samples with highest scores over all classes are removed, where $\beta$ is the total number of detection steps and $\epsilon$ is the fraction of samples to be ultimately removed (after $\beta$ steps); the classifier is then retrained on the remaining samples. The performance of their detector sensitively depends on the choice of the hyperparameters $\beta$ and $\epsilon$, as well as on the chosen training loss function. 
Note that there is no {\it a priori} knowledge of a good choice for $\epsilon$.
Also, their method is only applicable to linear classifiers, \eg, an SVM. Besides, they only report the improvement in the test set error rate -- the false positive rate is not mentioned, even though this is an important criterion for assessing a defense against DP attacks. 
\cite{DiakonikolasKK019} is computationally expensive as it requires performing an SVD for each class, at each detection step, and retraining the classifier after each detection step. 

\cite{Hong20GS} mitigates the effects of DP by gradient shaping (GS), \ie, constraining the magnitude and orientation of poisoned gradients to make them close to clean gradients. For example, one can adopt a differentially-private mechanism based optimizer (\eg differentially-private stochastic gradient descent (DP-SGD)) in training. A DP-SGD optimizer clips gradients according to the hyper-parameter \textit{$l_2$\_norm\_clip} and then adds noise to the gradients, whose size is controlled by the hyper-parameter \textit{noise\_multiplier}.
% Both \cite{DiakonikolasKK019} and \cite{Hong20GS} are designed against several types of DP attacks and applicable to various machine learning tasks. 
\cite{Hong20GS} is computationally cheap -- it does not require extra computation pre-training/post-training. However, their method only reduces the effect of poisoning, rather than eliminating the poisoned samples. Efficacy of their defense is dramatically degraded as more and more attack samples are injected, as will be seen from our results.

\cite{mixture} applied a BIC-based defense against DP attacks for binary classification tasks. The fundamental difference between \cite{mixture} and our work pertains to \textit{untainted data availability}. 
% Besides, our method can be applied to cases with more than two classes.
Their method assumes that the attacker only poisons one of the two classes, with this class known to the defender. 
Thus, the defender can always take the clean class as reference
in helping to identify poisoned samples in the corrupted class (This should be especially helpful for label flipping attacks, where the poisoned samples, labeled to one class, will be typical of the other (clean) class). However, in practice, the attacker is able to poison more than one class, and the defender does not know which class(es) are poisoned. 
Under this most realistic attack scenario, \cite{mixture} may fail even if only one class is poisoned, as the defender might sanitize the clean class based on the poisoned one (\cf Section \ref{experiments_bi}).

\cite{PaudiceML18, Hong20GS, DiakonikolasKK019} are supervised detection methods, with their performances highly impacted by the choices of hyper-parameters.
By contrast, our method is unsupervised.
At each optimization step, it separately assesses the hypothesis that each individual mixture component, in each class, is poisoned.
% one of the classes (and one of the mixture components representing this class) 
Only the component whose trial-sanitization yields the lowest BIC cost is actually sanitized. This process is repeated until the total BIC cost, defined over all classes, converges.

\section{Threat Model}\label{threat model}

In this paper, we consider $W$-class ($W\ge 2$) classification tasks, where the classifier, denoted 
$$f: \mathbb{R}^d \rightarrow \{1,\dots,W\},$$
is trained on $\mathcal{D}_{\text{Train}}$ and then tested on $\mathcal{D}_{\text{Test}}$. $\mathcal{D}_{\text{Train}}$ and $\mathcal{D}_{\text{Test}}$ are assumed i.i.d. from the same (unknown) distribution. Each feature $x_l, ~l=1,\ldots,d$, may be either discrete or continuous-valued. Both continuous and discrete-valued feature spaces will be considered in our experimental results. 

We assume the attacker: 1) has sufficient knowledge of the classification domain to generate or acquire samples that are legitimate instances of the different classes; 2) has access to covertly insert poisoned samples into the training set ($\mathcal{D}_{\text{Train}} = \mathcal{D}_{\text{Clean}} \cup \mathcal{D}_{\text{Attack}}$); 3) May simultaneously poison any subset of the classes, possibly with different attack strengths (\ie, different amounts of poisoned samples) for individual classes; 4) is unaware of any deployed defense. The \textit{goal} of the attacker is to degrade the classifier's (test set) generalization accuracy as much as possible, {\it i.e.},
\begin{equation*}
P_c = \frac{1}{|\mathcal{D}_{\text{Test}}|} \sum_{\boldsymbol{x}_i\in\mathcal{D}_{\text{Test}}}\mathbbm{1}(f(\boldsymbol{x}_i), y_i), 
\end{equation*}
where $\mathbbm{1}(a,b)$ is an indicator function with value 1 when $a=b$ and value 0 otherwise.

We assume the defender: 1) can only use the training set $\mathcal{D}_{\text{Train}}$ manipulated by the attacker, not any additional samples known to be clean (attack-free) -- this is in line with the embedded DP scenario; 2) does not know whether an attack is present, and if so, does not know the subset of attacking samples ($\mathcal{D}_{\text{Attack}}$), nor which class(es) are corrupted. The defender \textit{aims} to: 1) identify and remove as many poisoned samples as possible and as few clean samples as possible, before classifier training/retraining, and in so doing: 2) maintain classification accuracy as close as possible to that of a classifier trained on a clean (unpoisoned) data set.

\section{BIC Mixture-based Sanitization Strategy}\label{method}

We apply a data sanitization strategy on the training set, \ie, identifying and removing poisoned samples as training set outliers prior to classifier training. To accurately describe the possibly poisoned dataset, we apply mixture modeling to each class. Mixture modeling is a sound statistical approach for well-fitting potentially multi-modal data \cite{Duda,Mclachlan} and also gives the potential for concentrating the poisoned samples into just a few components, which assists in accurately identifying and removing them. Note that, in practice, poisoned components may own both poisoned and untainted samples, with the poisoning ratio for each component unknown. 

An attacker can confound the learning of class-discriminating features by choosing, for poisoning, samples typical of one class, but then labeling them to other class(es) ({\it i.e.}, a label flipping attack).  {\it Accordingly, we would expect that such samples are better explained by the mixture model for a class other than the class to which they are labeled.}
Thus, we propose to identify poisoned samples in the training set as those with greater {\it likelihood} under a class different from the one to which they are labeled.  We effectively {\it re-assign} such samples to the class (and mixture component) under which they have the greatest likelihood.

Now, suppose the vast majority of a mixture component's samples are re-assigned in this way to another class.  In this case, there may be insufficient remaining samples to reliably (or even in a well-posed fashion\footnote{For example, for a multivariate Gaussian component, with a full covariance matrix of size $d\times d$, one needs at least $d$ samples to estimate a (full-rank) covariance matrix.}) estimate the component's parameters.  In such a case, rather than retaining this component, it may be better 
to {\it remove} it, with its remaining samples assigned to other components in the class's mixture model.
A principled, theoretically supported criterion 
for model order selection, suitable for
use in deciding between revising and (wholesale) removing a mixture component from a class's model, is the {\it Bayesian Information Criterion} (BIC), which expresses an inherent tradeoff between data likelihood fit and model complexity.
The BIC objective function for a given data set $\cal{D}$ is defined as:
\begin{eqnarray}
\label{BIC}
{\rm BIC} = |\theta|k - L({\cal D}; \theta),
\end{eqnarray}
where $\theta$ is the set of free parameters specifying a density function model for the data, $|\theta|$ is the number of free parameters in this set, $k$ is the cost (penalty) for describing an individual model parameter, and $L(\mathcal{D}; \theta)$ is the log-likelihood of the data set $\mathcal{D}$, based on the density function model.  
In \cite{schwarz1978}, under suitable assumptions, BIC is 
shown to be a consistent estimator of model order.
In \cite{lanterman2001}, 
within an approximate Bayesian setting, the BIC cost for describing an individual model parameter is derived, and found to be $k=0.5*\log(|\mathcal{D}|)$.  
This model penalty will be used in the following.

The form of the BIC objective seen above is equivalent to the minimum description length (MDL) \cite{MDL},
and is amenable to interpretation as a two-part codelength: i) the first term, which we will denote by 
$\Omega$ in the sequel, is the number of bits needed to describe the parameters of the density model for the data; ii) the second, negative log-likelihood term is the number of bits to describe the data set, given the model.

In this work, we model the training data labeled to each class by a  class-specific mixture of density functions (or probability mass functions in the case of discrete data), \ie,
for an individual sample $\boldsymbol{x}$, labeled to class $c$, its density (likelihood) is: 
$$P[\boldsymbol{x};\theta_c] = \sum\limits_{j=1}^{M_c} \alpha_{j}^c 
%P_{j}^c
P[\boldsymbol{x};\Lambda_{j}^c],$$
where $\alpha_{j}^c$ is the probability mass (prior probability) of mixture component $j$ for class $c$ (\ie, $\sum\limits_{j=1}^{M_c} \alpha_{j}^c=1$, $\alpha_{j}^c\geq 0$ $\forall j$), 
%$P_{j}^c[\cdot]$ 
$P[\cdot ;\Lambda^c_j]$ 
is the $j^{\rm th}$ mixture component density under class $c$, $\Lambda_{j}^c$ is the set of parameters specifying the component density, and $M_c$ is the number of mixture components in the density for class $c$.  Note that 
$\theta_{c} = \{\Lambda_{j}^c\}\bigcup\{\alpha_{j}^c\}$ and $\theta=\bigcup_c \theta_{c}.$ 

In the above, a data sample from class $c$ is associated probabilistically with all mixture components from the class's mixture density.  Alternatively,
in this work, we consider the {\it complete data} BIC objective function, based on the complete data log-likelihood function \cite{Dempster}, wherein each data sample is hard (fully) assigned to the mixture component under which it has the greatest log-likelihood. That is, the complete data log-likelihood for the data from 
class $c$ is\footnote{Here we assume the component priors are uniform and hence
they are absent from the complete data log likelihood. In practice, these
terms do not affect detection performance significantly.} 
$$L_j^c = \sum\limits_{\boldsymbol{x} \in {\cal X}_{j}^c} \log P[\boldsymbol{x}; \Lambda_{j}^c], $$
where $\boldsymbol{x} \in {\cal X}_{j}^c$ if and only if, for 
$\boldsymbol{x}$ labeled to class $c$, $P[\boldsymbol{x} ; \Lambda_{j}^c] \geq P[\boldsymbol{x};\Lambda_{j'}^c]$  $\forall j' \neq j$.

Likewise the complete data BIC objective is:
\begin{eqnarray}
\label{completeBIC}
{\rm BIC}_{\rm cmplt} = |\theta|k - \sum\limits_{c=1}^W\sum\limits_{j=1}^{M_c} L_j^c. 
\end{eqnarray}

In short, the principle behind our complete data BIC-based defense is: \textit{a component is identified as poisoned if removing or revising it and re-assigning its samples reduces the BIC cost;  moreover, samples which are redistributed to other class(es) are deemed poisoned.} Thus, our anomaly detection method is \textit{unsupervised} and consistent with solving a BIC minimization problem\footnote{Apart from poisoned samples, our method might also remove any outliers, if it is BIC-efficacious to do so.}. 
In the sequel, we develop an algorithm for mitigating data poisoning via (locally optimal) minimization of the above complete data BIC objective.

\subsection{BIC-based Defense}

Recall $\{1,\dots,W\}, W\ge2$, is the set of classes, and let
$T=|\mathcal{D}_{\text{Train}}|$ be the total number of training samples.  Let $(\boldsymbol{x}_i,y_i)\in \mathbb{R}^d \times \{1,\dots,W\}$ represent the feature vector and label of training sample $i$. Denote $\Omega$ and $L$ as the model complexity and data log-likelihood, respectively. $M^c$ is the number of components in class $c$, with each specifying, \eg, a joint probability mass function (PMF) or probability density function (PDF) for component $j$ of class $c$, depending on whether the data is discrete or continuous-valued, respectively. 
%All components have the same number of parameters which %is denoted as $N_{\theta}$.

The model parameters $\theta_c$ of the mixture for class $c$ are estimated via the Expectation-Maximization (EM) algorithm \cite{Dempster}, applied to the subset of 
${\cal D}_{\rm train}$ labeled as class $c$.
The chosen model order $M^c$ is the one that yields the least BIC cost \eqref{BIC} over the set $\{1,\ldots,M^c_{\max}\}$ \cite{schwarz1978}, with $M^c_{\max}$ an upper bound on the number of components in class $c$'s mixture\footnote{$M^c_{\max}$ is in fact not a hyperparameter, as one can observe the changes of BIC to adjust the range of model orders. For example, if $M^c_{\max}$ yields the least BIC, one can increase $M^c_{\max}$ and repeat model selection until $M^c\neq M^c_{\max}$.}. 
Finally, we let $\mathcal{S}=\{(c,j)|c=1,\dots,W, j=1,\dots,M^c\}$ be the set of components across all classes. 

As just mentioned, we first estimate a mixture model for each class and then identify poisoned components and samples by minimizing the complete data BIC cost 
\eqref{completeBIC}
defined in the last section. 
This BIC minimization involves sample re-distribution, component removal/revision, and parameter updates. To reflect these model changes, we introduce several types of ``indicator'' variables:
\begin{itemize}
    \item     The class $t_i$ and component under this class $j_i$ that best-explain sample $\boldsymbol{x}_i$ are
    $$(t_i,j_i)=\argmax_{t=\{1,\dots,W\},~j=\{1,\dots,M^c\}} P[\boldsymbol{x}_i;\Lambda_j^t],$$ 

    \item $r_j^c=\left\{
    \begin{aligned}
    &1 \quad \text{component } j \text{ in class } c \text{ is poisoned} \\
    &0 \quad \ow
    \end{aligned}
    \right.
    $
    \item $q_j^c=\left\{
    \begin{aligned}
    &1 \quad \text{component } j \text{ in class } c \text{ needs to be revised} \\
    &0 \quad \text{component } j \text{ in class } c \text{ needs to be removed}
    \end{aligned}
    \right.
    $
\end{itemize}
Note that $q_j^c$ is configured only when $r_j^c=1$.

To account for possible data poisoning in this paper,
the complete data BIC cost to be minimized is 

\begin{flalign}\label{newbic}
    \nonumber
    &{\rm BIC}_{\rm cmplt}(\theta)  =  \sum\limits_{c=1}^W\sum\limits_{j=1}^{M_c}((1-r_j^c(1-q_j^c))k|\Lambda_j^c|+1 + \delta(r_j^c,1))& \\
    &- \sum\limits_{c=1}^W\sum\limits_{j=1}^{M_c}((1-r_j^c)L_j^c(\Lambda_j^c) + r_j^c\sum\limits_{\boldsymbol{x}_i \in {\cal X}_j^c} \log P[\boldsymbol{x}_i; \Lambda_{j_i}^{t_i}]).&
\end{flalign}

% \begin{eqnarray}\label{newbic}
% \nonumber
% {\rm BIC}_{\rm cmplt}(\theta)  =  \sum\limits_{c=1}^W\sum\limits_{j=1}^{M_c}((1-r_j^c(1-q_j^c))k|\Lambda_j^c|+1 + \delta(r_j^c,1)) \\
% - \sum\limits_{c=1}^W\sum\limits_{j=1}^{M_c}((1-r_j^c)L_j^c(\Lambda_j^c) + r_j^c\sum\limits_{\boldsymbol{x}_i \in {\cal X}_j^c} \log P[\boldsymbol{x}_i; \Lambda_{j_i}^{t_i}]).
% \end{eqnarray}
In \eqref{newbic}, the model parameters are
$\theta=\{\{\Lambda_j^c\},\{r_j^c\},\{q_j^c\}\}$, where the structural parameters
$r_j^c$ and $q_j^c$  each require one bit to specify (hence the `1' and $\delta(r_j^c,1)$ contributions to the model complexity term). By contrast, $t_i$ and $j_i$ are  hidden data assignments (as part of the complete data log-likelihood, and complete data BIC) not model parameters.

To minimize \eqref{newbic} in a locally optimal fashion,
our approach will involve cycling over the mixture components, one at a time, effecting changes to the mixture models that reduce this objective.   
The new BIC cost, in light of changes to component $j$ from class $c$,
can be expressed as the ``old'' BIC cost plus the (negative) change resulting from sample re-assignments or component removal, denoted
$\Delta {\rm BIC}_j^c $.

Each feasible joint configuration of the variables for component $j$ in class $c$ corresponds to one of three cases:
% The total full BIC cost is 
%     $$BIC = \sum_{(k,j')\in \mathcal{S}}\frac{1}{2}N_{\theta}\log T - \sum_{(k,j')\in \mathcal{S}}\sum_{\boldsymbol{x}_i\in \mathcal{X}_{j'}^k}\log P[\boldsymbol{x}_i ;\theta_{j'}^k]$$

1) $r_j^c=0$: The component is formed by clean samples, and there is no need to re-distribute its samples or modify the component (\ie, $\Delta\Omega^c_{j,1}=0$, $\Delta L^c_{j,1}=0$).
The change in BIC in this case is thus
$$\Delta {\rm BIC}_j^c = \Delta\Omega^c_{j,1} + \Delta L^c_{j,1}  = 0.$$
    
2) $r_j^c=1, q_j^c=0$: Component $j$ is poisoned, and we are choosing to remove it from the mixture, changing the model complexity term by 
$$\Delta\Omega^c_{j,2} = -|\Lambda_j^c|\frac{1}{2}\log T,$$
where $|\Lambda_j^c|$ is the number of parameters in component $j$ from class $c$.
The component's samples are re-distributed consistent with maximizing the log-likelihood: Each sample $\boldsymbol{x}_i \in \mathcal{X}_j^c$ is re-assigned to component $j_i$ of class $t_i$, where
$$(t_i, j_i)=\argmax_{(t,j')\in\mathcal{S}\setminus\{(c,j)\}}\log P[\boldsymbol{x}_i;\Lambda_{j'}^t].$$
Let $$\mathcal{Q}=\{(t_i, j_i) | \forall i,~ \boldsymbol{x}_i \in \mathcal{X}_j^c\}$$ be the set of components which receive the re-assigned samples.  
For each component $(w, j')\in\mathcal{Q}$, we re-estimate its parameters on $\widehat{\mathcal{X}_{j'}^w}$ by maximum likelihood estimation (MLE):
$$\Lambda_{j'}^{w,\text{new}} = \argmax_{\Lambda}\sum_{\boldsymbol{x}_i\in \widehat{\mathcal{X}_{j'}^w}} \log P[\boldsymbol{x}_i;\Lambda],$$ where
$$\widehat{\mathcal{X}_{j'}^w} = \mathcal{X}_{j'}^w \cup \{\boldsymbol{x}_i \in \mathcal{X}_j^c | t_i = w, j_i = j' \}.$$
This optimization has a closed form, globally optimal solution for the component density model forms considered in this paper.
The total data log-likelihood changes by
\begin{equation*}
    \begin{aligned}
        \Delta L_{j,2}^c &=  -\sum_{(w,j')\in\mathcal{Q}}\sum_{\boldsymbol{x}_i\in\widehat{\mathcal{X}_{j'}^w}}\log P[\boldsymbol{x}_i;\Lambda_{j'}^{w,\text{new}}] \\
        & + \sum_{(w,j')\in\mathcal{Q}}\sum_{\boldsymbol{x}_i\in\mathcal{X}_{j'}^w}\log P[\boldsymbol{x}_i;\Lambda_{j'}^w] + \sum_{\boldsymbol{x}_i\in\mathcal{X}_j^c}\log P[\boldsymbol{x}_i;\Lambda_j^c].
    \end{aligned}
\end{equation*}
The change in BIC in this case is 
$$\Delta {\rm BIC}_j^c = \Delta\Omega^c_{j,2} + \Delta L^c_{j,2}.$$

3) $r_j^c=1, q_j^c=1$: Similar to case (2) but instead of removing it, we re-estimate the parameters of component $j$ by its surviving samples (\ie, samples with $t_i=c$). Revising a component does not change the model complexity cost, since the code length is untouched (\ie, $\Delta\Omega^c_{j,3}=0$).
The parameters $\Lambda_j^{c}$ are re-estimated by MLE on the surviving samples:
$$\Lambda_j^{c,new} = \argmax_{\Lambda} \sum_{\boldsymbol{x}_i\in \widehat{\mathcal{X}_j^c}} \log P[\boldsymbol{x}_i ; \Lambda],$$ where
$$\widehat{\mathcal{X}_j^c} = \{\boldsymbol{x}_i\in \mathcal{X}_j^c | t_i = c\}.$$
Samples that are best represented by class $w\neq c$ (\ie, $t_i=w$, $w\neq c$) are re-distributed to their fittest components in class $w$, but the remaining samples (\ie, $t_i=c$) are explained by the updated component $j$.
Let
$$\mathcal{Q}' = \{(w,j')\in\mathcal{Q}|w\neq c\}\cup\{(c,j)\}$$
be the set of components to be updated. 
The total data log-likelihood changes by
\begin{equation*}
    \begin{aligned}
        \Delta L_{j,3}^c = & - \sum_{(w,j')\in\mathcal{Q}'}\sum_{\boldsymbol{x}_i\in \widehat{\mathcal{X}_{j'}^w}} \log P[\boldsymbol{x}_i;\Lambda_{j'}^{w,\text{new}}] \\
        & + \sum_{(w,j')\in\mathcal{Q}'}\sum_{\boldsymbol{x}_i\in \mathcal{X}_{j'}^w} \log P[\boldsymbol{x}_i;\Lambda_{j'}^w],
    \end{aligned}
\end{equation*}
where $\widehat{\mathcal{X}_{j'}^w}$ and $\Lambda_{j'}^{w,\text{new}}$ $\forall (w,j')\in\mathcal{Q}'\setminus\{(c,j)\}$ are defined in the same way as in case 2.
The BIC change in this case is
$$\Delta {\rm BIC}_j^c = \Delta L^c_{j,3}.$$

%Adding the BIC contributions from all components in all %classes, considering all three cases, we have the total %BIC objective function:
%\begin{equation}\label{BIC}
%\begin{aligned}
%    BIC \propto & \sum_{(c,j)\in\mathcal{S}} [
%       r_j^c (1-q_j^c) (\Delta\Omega^c_{j,2} + \Delta %L^c_{j,2}) + r_j^c q_j^c \Delta L^c_{j,3}
%    ]
%\end{aligned}
%\end{equation}

To minimize the complete data BIC objective, 
for each component in class $c \in\{1,\dots,W\}$, 
we should choose the configuration of the parameters that reduces BIC the most. However, the optimal configuration for any component $j$ depends on the configurations for other components. It is thus intractable to define an algorithm guaranteed to find a globally optimal configuration over all components ({\it e.g.}, by exhaustively evaluating over the huge combinatorial space of component configurations). 
Instead, at each optimization step, we separately {\it trial}-update each component's configuration, and then only permanently update for the component that yields the greatest reduction in BIC.  This is repeated until there are no further changes. This optimization approach is non-increasing in the BIC objective and results in a locally optimal solution.

The null hypothesis of our detection inference is that the training set is not poisoned (and is generated according to the existing mixture model).
If there is data poisoning, the training set is hypothesized to be generated by an alternative model (with some components removed and/or modified). Thus, we perform the following hypothesis testing: after BIC minimization, if $r_j^c = 0$ holds for all components in all classes, and $t_i = y_i$
holds for all samples, then no components and samples are inferred to be poisoned, and we accept the null hypothesis. Otherwise, we reject the null hypothesis, and the training set is deemed poisoned.
The samples that were re-assigned to other classes via the BIC minimization are deemed poisoned, and are removed from the training set.

\subsection{Implementation}
Consistent with the above description, we apply an iterative, locally optimal approach to minimize the total BIC cost and optimize its parameters, as shown in Algorithm~\ref{pseudocode}. 
As we discussed before, at each algorithm step, we only sanitize the component whose removal/revision decreases the total BIC cost the most. 
That is, we first evaluate the reduction in the total BIC cost $\Delta {\rm BIC}^c_j$ caused by trial removal/revision of each component $j$ in each class $c$.
Then, we sanitize the component $j^*$ in class $c^*$ which decreases the total BIC cost the most, \ie, 
$$(c^*,j^*) =\argmin_{c\in\{1,\dots,W\},j=1,\dots,M^{c}}\Delta {\rm BIC}^{c}_j,$$
where 
\begin{align*}
    \Delta {\rm BIC}^{c}_j &= \min_{m=1,2,3} \{\Delta\Omega^c_{j,m}+\Delta L^c_{j,m}\}.
                    %  &= \min_{m=1,2,3} \{\Delta\Omega^c_{j,m} - (L^c_{j,m}-L^c_{j,1}) \}
\end{align*}
The above procedure is repeated until the total BIC cost converges, \ie until no trial component updates further reduce the BIC cost. Finally, all samples with $t_i\neq y_i$
(\ie, the detected poisoned samples) are removed from the training set, and we have the sanitized training set $$\widehat{\mathcal{D}}_{\text{Train}}=\{\boldsymbol{x}_i\in\mathcal{D}_{\text{Train}}|t_i=y_i\}.$$

\begin{algorithm}[ht]
\SetAlgoLined
\SetKwInOut{Input}{Input}
\SetKwInOut{Output}{Output}
\SetKwRepeat{Do}{do}{while}
\Input{$\mathcal{D}_{\text{Train}}=\{(\boldsymbol{x}_i, y_i)\}_{i=1}^N, \{\{\Lambda_j^c\}_{j=1}^{M^c}\}_{c\in\{1,\dots,W\}}$}
% \Output{$\{\{r_j^c\}_{j=1}^{M^c},\{q_j^c\}_{j=1}^{M^c}\}_{c\in\{1,\dots,W\}}$, $\{t_i\}_{i=1}^N$}
\Output{$\widehat{\mathcal{D}}_{\text{Train}}$}
$r_j^c=0$, $q_j^c=0$, $\forall c,j$ \;
$t_i = y_i$, $\forall i$\;
  $\Delta {\rm BIC}^c_j = 0, \forall j,c$\;
 \Do{$\sum_{c,j}\Delta {\rm BIC}^c_j < 0$}{
  \For{$c\in\{1,\dots,W\}$}{
    \For{each component $j$ in class $c$}{
        compute BIC reduction from $j$ for the three cases: $\{\Delta\Omega^c_{j,m}+\Delta L^c_{j,m}\},m=1,2,3$ \;
configure $\{t_i ~|~ \boldsymbol{x}_i\in \mathcal{X}_j^c$, $r_j^c$, $q_j^c$ to $\min\{\Delta\Omega^c_{j,m}+\Delta L^c_{j,m}\}_{m=1}^3$\;
        $\Delta {\rm BIC}^c_j=\min\{\Delta\Omega^c_{j,m}+\Delta L^c_{j,m}\}_{m=1}^3$\;
        }
    }
    $(c^*,j^*)=\argmin_{c\in\{1,\dots,W\},j=1,\dots,M^{c}}\Delta {\rm BIC}^{c}_j$\;
    \If{$r_{j^*}^{c^*}=1$}{
        For $\boldsymbol{x}_i\in \mathcal{X}_{j^*}^{c^*}$, if $t_i=w, w\neq c^*$, re-distribute it to component $m=\argmax\limits_{m'}\log P[\boldsymbol{x}_i ;\Lambda_{m'}^w]$ in class $w$ and then update component $m$'s parameters via MLE\;
        
         \eIf{$q_{j^*}^{c^*}=0$}{
            remove component ${j^*}$ from $\{\Lambda_j^{c^*}\}_{j=1,\dots,M^{c^*}}$\;
            re-distribute each $\boldsymbol{x}_i\in \mathcal{X}_{j^*}^{c^*}$ to component $m=\argmax\limits_{m'}P[\boldsymbol{x}_i ;\Lambda_{m'}^{c^*}]$ and update component $m$'s parameters\;
         }{
            update component ${j^*}$'s parameters on $\mathcal{X}_{j^*}^{c^*}$\;
         }
    }
 }
 $\widehat{\mathcal{D}}_{\text{Train}}=\{\boldsymbol{x}_i\in\mathcal{D}_{\text{Train}}|t_i=y_i\}$\;
 \caption{BIC-Based Defense Against DP Attacks}\label{pseudocode}
\end{algorithm}
Note that: 1) the same component may be re-optimized multiple times during the course of this algorithm; 2) But removal of a component is permanent, \ie once removed, a component cannot be reinstated.
\section{Experiments on Binary Classification Tasks}\label{experiments_bi}

\subsection{Experiment Setup}\label{exp_setup}
\textbf{Dataset and mixture model}: 
% To demonstrate the effectiveness of our defense, we respectively apply it on a binary classification task and a multi-class classification task.
For binary ($W=2$) classification, we use the TREC 2005 spam corpus (TREC05) \cite{TREC05}. TREC05 contains 39,999 real ham and 52,790 spam emails which are labeled based on the sender/receiver relationship. For training, we randomly select 9000 ham emails and 9000 spam emails\footnote{There are 8651 ham emails and 8835 spam emails remaining in the training set after pre-processing. Some emails are removed since there are no tokens left after \eg, stop word removal and low-frequency word filtering.}. 
For testing, we randomly select 3000 ham emails and 3000 spam emails. The remaining samples are used for poisoning. The dictionary, following case normalization, stop word removal, stemming and low-frequency word filtering, has about 30,000 unique words. 

\begin{figure}
    \centering
    \includegraphics[width=0.4\textwidth]{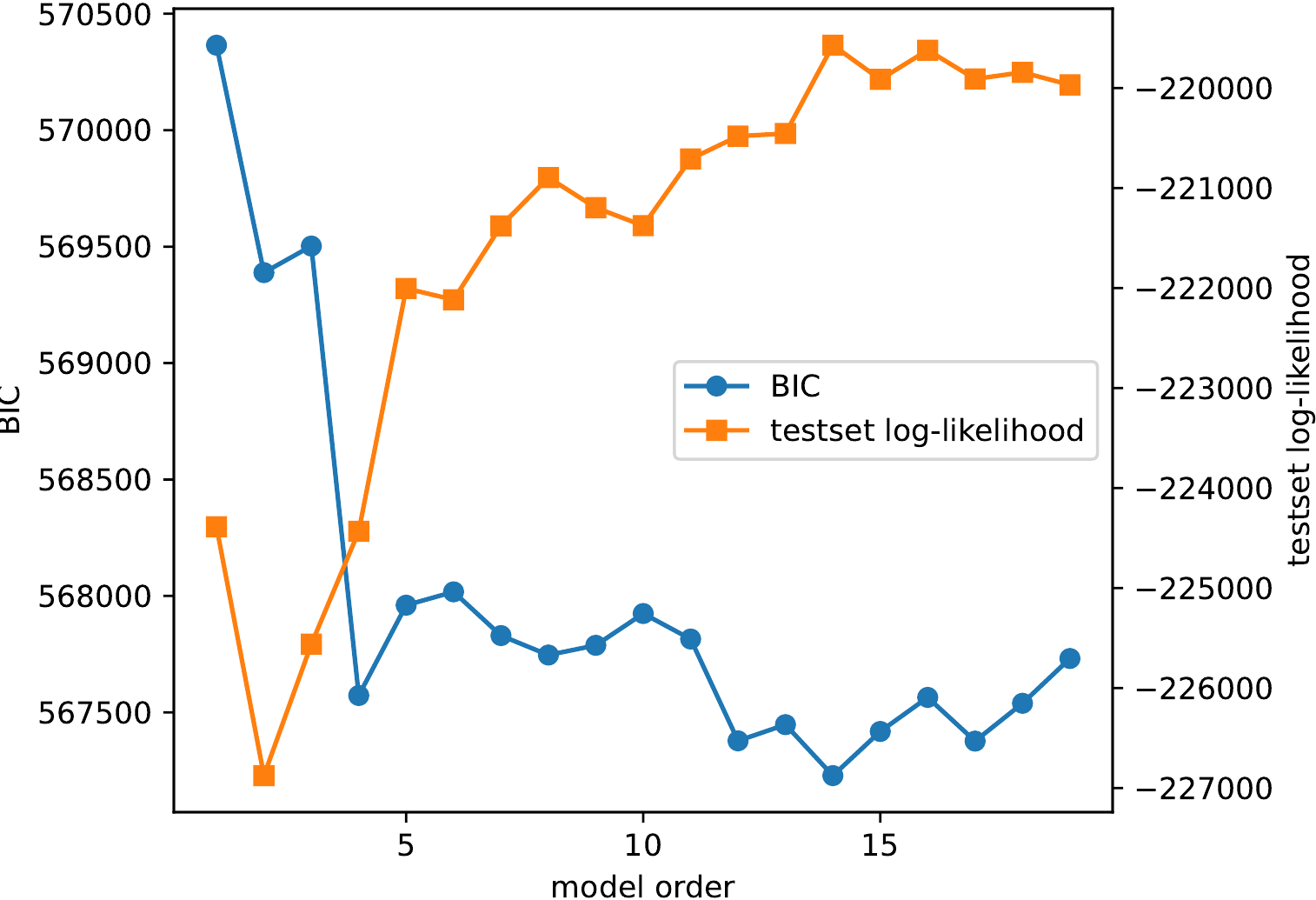}
    \caption{Training set BIC cost and test set log-likelihood as a function of the model order of the PMM on class ``soc.religion.christian'' of clean 20-Newsgroups dataset.}
    \label{fig:BIC-PMM}
\end{figure}

We apply Parsimonious Mixture Modeling (PMM) \cite{GrahamMiller06} on both datasets. 
PMMs allow parameter sharing across multiple components, which greatly reduces the number of model parameters compared with standard (unstructured) mixtures, and which allows BIC to choose good model orders in high feature dimensions, rather than grossly underestimating the model order (number of mixture components).
Figure \ref{fig:BIC-PMM} shows that, for class ``soc.religion.christian'' of the clean 20-Newsgroups dataset, PMM chooses a reasonable model order (14)  that minimizes the training set BIC cost and also has good generalization (\ie, test set log-likelihood).
PMMs' superiority over standard unstructured mixture models for high dimensional datasets such as text was demonstrated in  \cite{GrahamMiller06}. Initially we chose $M^c_{\max}=25$, $\forall c$. If the chosen model order $M^{c}=M^c_{\max}$, then $M^c_{\max}$ is increased and the model is retrained. For both datasets, the training and test samples are represented using a bag-of-words. Each PMM component is a multinomial joint probability mass function.

\textbf{DP attack and target classifiers}: 
To simulate a reasonable and potent embedded data poisoning attack, we used real samples as attacking samples. That is, we inject samples from class $c$ into class $w\neq c$ as poisoned samples, with these samples intentionally mislabeled to class $w$. The attack strength, \ie, the number of attacking samples injected, may not be the same for each class. The poisoned samples are randomly selected (from the sample pool left over from the training and test sets).

We launched 12 poisoning attacks on TREC05. For half of the attacks, we only poisoned one class (\eg, spam), with attack strength from 1000 to 6000 samples. For the other half of the attacks, we simultaneously poisoned the ham and spam training sets with various attack strengths (\cf Table\ref{tab:1}). 
% we launch a hybrid of ``pure-ham'' and ``pure-spam'' poisoning attacks on the training set, with real ham emails and spam emails injected into the clean spam and ham training sets, respectively.

% For both binary classification and multi-class classification tasks,
We chose linear SVM \cite{cortes1995support}, LR \cite{nelder1972generalized}, and bi-directional one-layer long short-term memory (LSTM) \cite{hochreiter1997long} recurrent neural networks with 128 hidden units as the target classifiers, since they are effective in the classification of text data. 

\textbf{Evaluation criteria}: 
The performance of our defense (BIC-D) was measured by: 1) improvement in test classification accuracy after data sanitization; 2) true positive rate (TPR) — the fraction of poisoned samples that are detected; and 3) false positive rate (FPR) – the fraction of non-poisoned samples falsely detected.
We also applied the BIC-based defense with clean data samples (BIC-C-D) \cite{mixture}, a KNN-based defense (KNN-D) \cite{PaudiceML18}, and a GS-based defense (GS-D) \cite{Hong20GS} on the same poisoned training sets and compared with them in terms of the above evaluation criteria. We failed to apply the SVD-based defense \cite{DiakonikolasKK019} on TREC05, since it is too expensive to perform SVD on the matrix of gradients, whose size is around (20000 x 60000).

\textbf{Hyperparameter setting:} 
1) For the KNN-based defense, we set the number of neighbors at $K=10$, 
% and the threshold of majority decision making at $\eta=0.5$,
which is suggested by \cite{PaudiceML18}. 2) For the GS-based defense, we applied a DP-SGD optimizer with clip-norm of 2.0 and noise multiplier of 0.1 for training -- these values were used in \cite{Hong20GS}. 
% 3)For the SVD-based defense, we set the number of detection steps $r=2$, which is the same with \cite{DiakonikolasKK019}. To show its best performance, we set $\epsilon$ as the real poisoning ratio if the training set is poisoned, and set it as 0.01 if there is no poisoning. 

\subsection{Experimental Results}

\newcommand{\tabincell}[2]{\begin{tabular}{@{}#1@{}}#2\end{tabular}} 

\begin{table*}[ht!] 
\centering
% \scriptsize
% \footnotesize
% \small
\begin{tabular}{p{2.5cm}|p{0.735cm}p{0.735cm}p{0.735cm}p{0.735cm}p{0.735cm}p{0.735cm}p{0.735cm}p{0.735cm}p{0.735cm}p{0.735cm}p{0.735cm}p{0.735cm}p{0.735cm}}
\toprule
\hline
\tabincell{c}{$\#$ Poisoned Ham,\\$\#$ Poisoned Spam} & 0, 0 & 0, 1000 & 0, 2000 & 0, 3000 & 0, 4000 & 0, 5000 & 0, 6000 & 1000, 1000 & 1000, 2000 & 2000, 1000 & 2000, 2000 & 2000, 4000 & 4000, 2000 \\
% \midrule
\hline
\multicolumn{13}{c}{\textbf{SVM}}\\
\hline
Poisoned & 0.9522 & 0.8867 & 0.8461 & 0.8215 & 0.7932 & 0.7731 & 0.7495 & 0.8339 & 0.7924 & 0.7833 & 0.7488 & 0.7142 & 0.7114 \\
BIC-D & \textbf{0.9684} & \textbf{0.9611} & \textbf{0.9530} & \textbf{0.9425} & \textbf{0.9411} & \textbf{0.9394} & \textbf{0.9329} & \textbf{0.9454} & \textbf{0.9284} & \textbf{0.9429} & \textbf{0.9143} & \textbf{0.8998} & \textbf{0.8731} \\
KNN-D & 0.9001 & 0.8974 & 0.8828 & 0.8660 & 0.8358 & 0.7958 & 0.7751 & 0.9049 & 0.8917 & 0.8880 & 0.8793 & 0.8421 & 0.8367 \\
GS-D & 0.9645 & 0.9372 & 0.9225 & 0.9023 & 0.8131 & 0.7042 & 0.6314 & 0.9129 & 0.8807 & 0.8738 & 0.8568 & 0.8159 & 0.7711 \\
BIC-C-D & 0.9579 & 0.9434 & 0.9124 & 0.8519 & 0.6882 & 0.6039 & 0.5697 & 0.9217 & 0.9088 & 0.9061 & 0.8288 & 0.6385 & 0.7153 \\
% SVD-D &  &  &  &  &  &  &  &  &  &  &  &  &  \\
\hline
\multicolumn{13}{c}{\textbf{LR}}\\
\hline
Poisoned & 0.9616 & 0.9175 & 0.8828 & 0.8443 & 0.8172 & 0.7803 & 0.7488 & 0.8843 & 0.8501 & 0.8481 & 0.8183 & 0.7591 & 0.7438 \\
BIC-D & \textbf{0.9699} & \textbf{0.9660} & \textbf{0.9559} & \textbf{0.9519} & \textbf{0.9461} & \textbf{0.9394} & \textbf{0.9368} & \textbf{0.9511} & \textbf{0.9402} & \textbf{0.9507} & \textbf{0.9315} & \textbf{0.9126} & \textbf{0.8799} \\
KNN-D & 0.9099 & 0.9073 & 0.8955 & 0.8831 & 0.8529 & 0.8069 & 0.7776 & 0.9169 & 0.9039 & 0.9044 & 0.8968 & 0.8646 & 0.8656 \\
GS-D & 0.9598 & 0.9384 & 0.9184 & 0.8606 & 0.8158 & 0.7099 & 0.6655 & 0.9258 & 0.9137 & 0.8948 & 0.8797 & 0.8091 & 0.7847 \\
BIC-C-D & 0.9622 & 0.9554 & 0.9247 & 0.8633 & 0.6909 & 0.6192 & 0.5801 & 0.9375 & 0.9222 & 0.9152 & 0.8358 & 0.6427 & 0.7158 \\
% SVD-D &  &  &  &  &  &  &  &  &  &  &  &  &  \\
\hline
\multicolumn{13}{c}{\textbf{LSTM}}\\
\hline
Poisoned & 0.9632 & 0.9363 & 0.9111 & 0.8852 & 0.8668 & 0.8159 & 0.8028 & 0.8788 & 0.8681 & 0.8691 & 0.8521 & 0.7738 & 0.7985 \\
BIC-D & \textbf{0.9701} & \textbf{0.9682} & \textbf{0.9619} & \textbf{0.9588} & \textbf{0.9513} & \textbf{0.9465} & \textbf{0.9424} & \textbf{0.9584} & \textbf{0.9476} & \textbf{0.9551} & \textbf{0.9431} & \textbf{0.9223} & \textbf{0.8991} \\
KNN-D & 0.9313 & 0.9281 & 0.9183 & 0.8941 & 0.8744 & 0.8449 & 0.8009 & 0.9317 & 0.9131 & 0.9013 & 0.9125 & 0.8905 & 0.8821 \\
GS-D & 0.8339 & 0.8205 & 0.8123 & 0.7792 & 0.7347 & 0.7153 & 0.6824 & 0.8383 & 0.8176 & 0.8198 & 0.8208 & 0.7718 & 0.7949 \\
BIC-C-D & 0.9629 & 0.9607 & 0.9217 & 0.8712 & 0.6915 & 0.6149 & 0.5906 & 0.9359 & 0.9232 & 0.9277 & 0.8404 & 0.6514 & 0.7368 \\
% SVD-D & - & - & - & - & - & - & - & - & - & - & - & - & - \\
\hline
\bottomrule
\end{tabular}
\caption{Test set classification accuracy of victim classifiers as a function of attack strength on poisoned and sanitized TREC05 datasets.}
\label{tab:1}
% \vspace{-6mm}
\end{table*}

\begin{table*}[!t] 
\centering
% \scriptsize
% \footnotesize
\begin{tabular}{p{2.5cm}|p{0.735cm}p{0.735cm}p{0.735cm}p{0.735cm}p{0.735cm}p{0.735cm}p{0.735cm}p{0.735cm}p{0.735cm}p{0.735cm}p{0.735cm}p{0.735cm}p{0.735cm}}
\toprule
\hline
\tabincell{c}{$\#$ Poisoned Ham,\\$\#$ Poisoned Spam} & 0,0 & 0, 1000 & 0, 2000 & 0, 3000 & 0, 4000 & 0, 5000 & 0, 6000 & 1000, 1000 & 1000, 2000 & 2000, 1000 & 2000, 2000 & 2000, 4000 & 4000, 2000 \\
% \midrule
\hline
\multicolumn{13}{c}{\textbf{True Positive Rates (TPRs)}}\\
\hline
BIC-D & - & \textbf{0.8898} & \textbf{0.9044} & \textbf{0.9036} & \textbf{0.8689} & \textbf{0.9014} & \textbf{0.8865} & 0.8633 & \textbf{0.8678} & 0.8874 & 0.8351 & 0.8113 & 0.8142 \\
KNN-D & - & 0.8393 & 0.8154 & 0.7856 & 0.7342 & 0.6478 & 0.5761 & \textbf{0.8996} & 0.8518 & \textbf{0.9082} & \textbf{0.8842} & \textbf{0.8362} & \textbf{0.8261} \\
% GS-based Defense & - & - & - & - & - & - & - & - & - & - & - & - & - \\
BIC-C-D & - & 0.8846 & 0.8340 & 0.7303 & 0.3644 & 0.1951 & 0.1122 & 0.8628 & 0.8420 & 0.8284 & 0.7446 & 0.2102 & 0.4390 \\
% SVD-D-SVM &  &  &  &  &  &  &  &  &  &  &  &  &  \\
% SVD-D-LR &  &  &  &  &  &  &  &  &  &  &  &  &  \\
\hline
\multicolumn{13}{c}{\textbf{False Positive Rates (FPRs)}}\\
\hline
BIC-D & \textbf{0.0177} & \textbf{0.0249} & \textbf{0.0841} & \textbf{0.0877} & \textbf{0.0553} & \textbf{0.0885} & \textbf{0.0652} & \textbf{0.0499} & \textbf{0.0629} & \textbf{0.0586} & \textbf{0.0737} & \textbf{0.0809} & \textbf{0.1131} \\
KNN-D & 0.0745 & 0.0826 & 0.0936 & 0.1095 & 0.1377 & 0.1798 & 0.2122 & 0.0888 & 0.1057 & 0.1012 & 0.1099 & 0.1339 & 0.1452 \\
% GS-based Defense & - & - & - & - & - & - & - & - & - & - & - & - & - \\
BIC-C-D & 0.0505 & 0.0724 & 0.0775 & 0.0881 & 0.3209 & 0.3621 & 0.3868 & 0.0598 & 0.0681 & 0.0630 & 0.2128 & 0.2958 & 0.2698 \\
% SVD-D-SVM &  &  &  &  &  &  &  &  &  &  &  &  &  \\
% SVD-D-LR &  &  &  &  &  &  &  &  &  &  &  &  &  \\
\hline
\bottomrule
\end{tabular}
\caption{TPRs and FPRs of three defenses on the TREC05 dataset under all attack cases.}
\label{tab:2}
% \vspace{-6mm}
\end{table*}

\begin{table*}[!t] 
\centering
% \scriptsize
% \footnotesize
\begin{tabular}{p{3.1cm}|p{0.7cm}p{0.7cm}p{0.7cm}p{0.7cm}p{0.7cm}p{0.7cm}p{0.7cm}p{0.7cm}p{0.7cm}p{0.7cm}p{0.7cm}p{0.7cm}p{0.7cm}}
\toprule
\hline
\tabincell{c}{$\#$ Poisoned Ham,\\$\#$ Poisoned Spam} & 0,0 & 0, 1000 & 0, 2000 & 0, 3000 & 0, 4000 & 0, 5000 & 0, 6000 & 1000, 1000 & 1000, 2000 & 2000, 1000 & 2000, 2000 & 2000, 4000 & 4000, 2000 \\
% \midrule
\hline
\# Components & (21,18) & (29,16) & (22,18) & (25,17) & (19,20) & (24,20) & (24,31) & (49,27) & (25,15) & (37,29) & (48,28) & (40,29) & (36,28) \\
\# Revised Components & (1,5) & (0,6) & (6,11) & (5,10) & (1,16) & (2,9) & (7,11) & (19,18) & (11,7) & (17,12) & (9,7) & (14,11) & (14,13) \\
\# Removed Components & (0,1) & (5,3) & (2,6) & (1,2) & (2,4) & (3,4) & (4,11) & (7,4) & (4,2) & (4,6) & (12,5) & (10,5) & (8,11) \\
\hline
\bottomrule
\end{tabular}
\caption{The number of components, number of revised components, and number of removed components of each class under all attack cases on the TREC05 dataset.}
\label{tab:3}
% \vspace{-6mm}
\end{table*}

\begin{figure}
    \centering
    \begin{subfigure}[b]{0.4\textwidth}
        \centering
        \includegraphics[width=\textwidth]{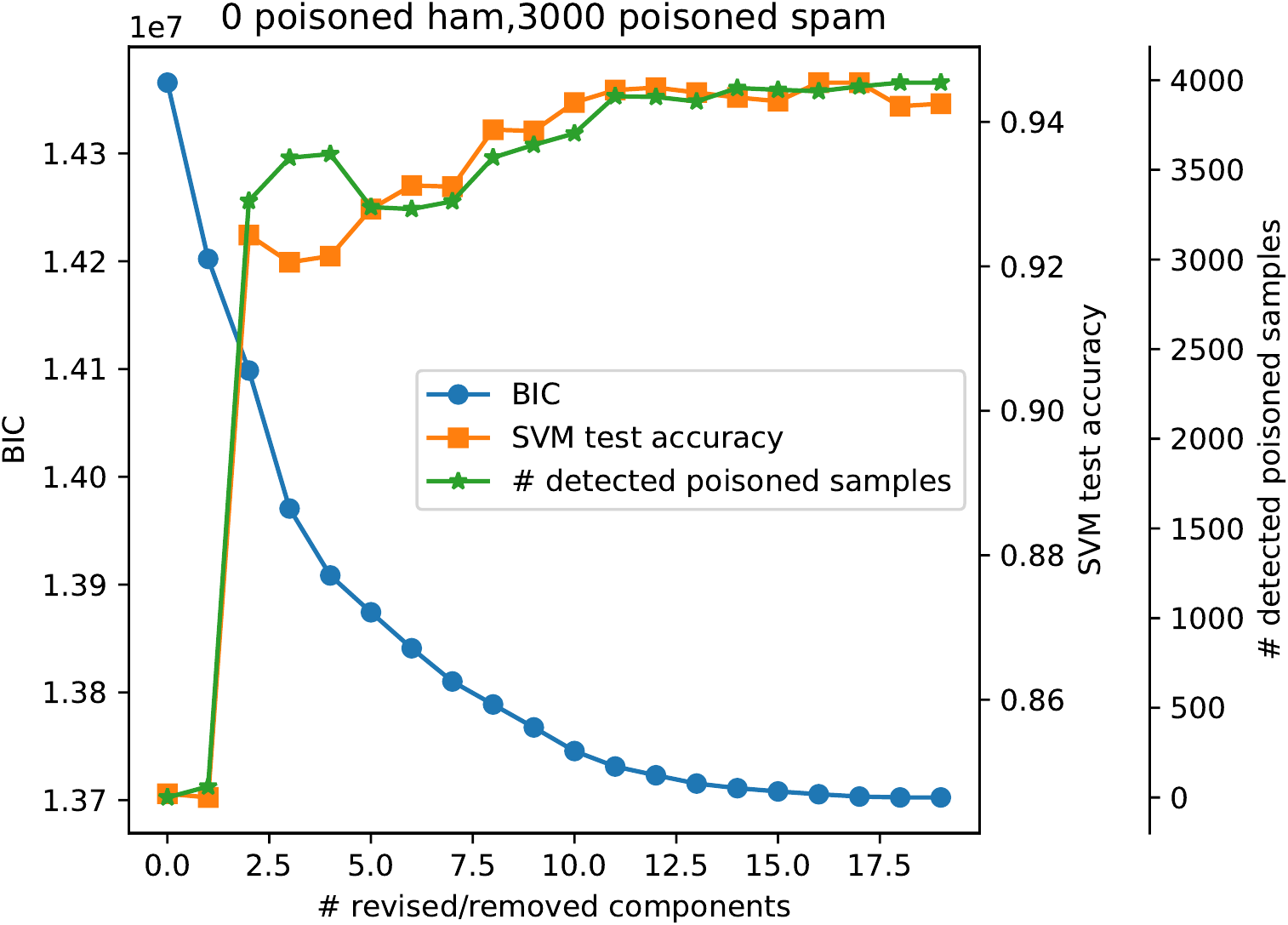}
        \caption{}
        \label{fig:spam-0-3000}
    \end{subfigure}
    \hfill
    \begin{subfigure}[b]{0.4\textwidth}
        \centering
        \includegraphics[width=\textwidth]{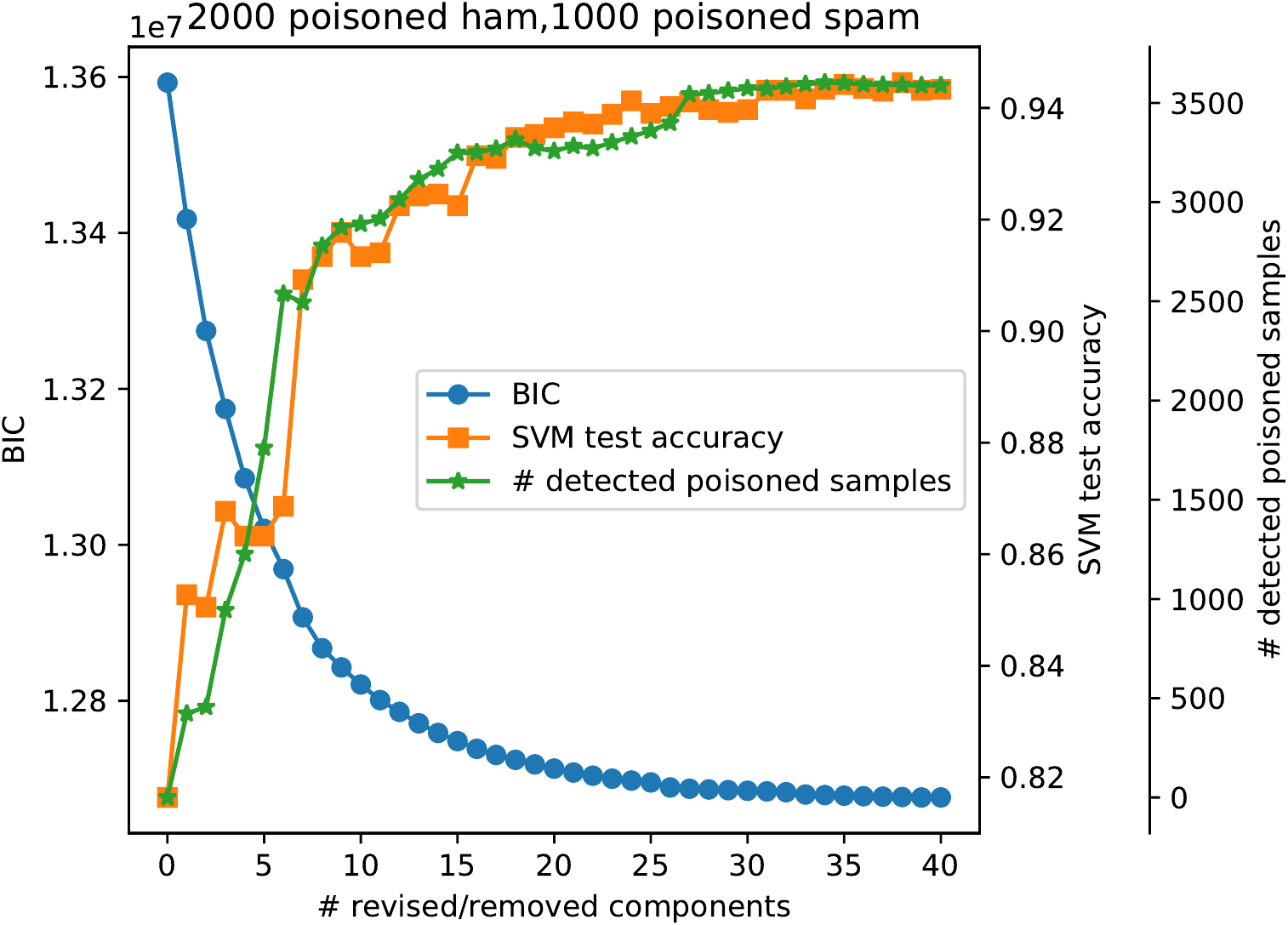}
        \caption{}
        \label{fig:spam-2000-1000}
    \end{subfigure}
    \caption{BIC cost, SVM test accuracy, and the number of detected poisoned samples as a function of the number of component change steps for attacks against TREC05 using (a) 0 poisoned ham samples, 3000 poisoned spam samples (b) 2000 poisoned ham samples, 1000 poisoned spam samples.}
    \label{fig:BIC-acc-vs-comp-spam}
\end{figure}

The results are listed in Table~\ref{tab:1} and \ref{tab:2}. Table~\ref{tab:1} shows the performance of victim classifiers as a function of attack strength on poisoned and sanitized TREC05 datasets. 
We first trained the target classifiers on the clean dataset (Attack (0,0) in Table \ref{tab:1}), yielding clean test accuracies (baselines) for SVM, LR, and LSTM of 0.9522, 0.9616, and 0.9632, respectively.
% Then we poison the training set with real emails and validate the effectiveness of the DP attack. 
The test accuracies of the classifiers poisoned by 12 DP attacks (described in Section \ref{exp_setup}) are shown as poisoned SVM/LR/LSTM in Table \ref{tab:1}.
As the total attack strength (the sum of the attacking ham and spam samples) is strengthened to 6000, the classification accuracies of SVM/LR drop below 0.75 and LSTM drops to 0.8. Thus, embedding real ham into the spam set and real spam into the ham set is indeed a significant poisoning attack on SVM/LR/LSTMs. 

Then, we applied the four defenses on the corrupted training sets and retrained the victim classifiers on the sanitized datasets. The corresponding test accuracies are shown as BIC-D, KNN-D, GS-D, and BIC-C-D in Table \ref{tab:1}.
Since BIC-C-D \cite{mixture} unrealistically assumes the defender knows which class is clean, we alternately apply BIC-C-D on ham and spam until the total BIC cost over the two classes converges. The order of sanitization was fixed -- it was always initiated from class ham. 
As expected, the test accuracies of the classifier with BIC-C-D drop rapidly when the total attacking strength exceeds 4000.
The test accuracies of classifiers using KNN/GS-based defenses also decline gradually as the attack strength increases, while our BIC-based defense performs well and stably. 
In all cases, our defense surpasses the other three defenses in classification accuracy (marked in bold). 
When the attack is strengthened to 5000 ham emails in the spam training set, the KNN-based defense exhibits little improvement in test accuracy over that of the poisoned classifier.
The classifiers equipped with the GS-based defense perform even worse than the poisoned classifiers. However, our defense significantly improves test accuracies even under strong attacks, restoring the test accuracies close to the clean baselines.
For LSTM with the GS-based defense, its accuracy is even worse than the poisoned LSTM in all cases. The performance of the GS-based defense is affected by the choice of its hyper-parameters -- clip norm and noise multiplier. We directly used the hyper-parameter settings from \cite{Hong20GS}, which are tuned for LR, not LSTM. 
See \cite{DBLP:conf/nips/BagdasaryanPS19, DBLP:conf/iclr/McMahanRT018} for tuning the hyperparameters of a DP-SGD optimizer.  Note that for the embedded data poisoning scenario considered here, there is no clean validation set available for (principled) tuning of hyperparameters.

Table \ref{tab:2} shows the TPRs and FPRs of the BIC/KNN-based defenses (since the GS-based defense does not identify the poisoned samples). Compared with KNN-D and BIC-C-D, our defense has relatively low FPRs for all cases. Besides, our defender has higher TPRs than the KNN-based defense when only the spam set is poisoned and has comparable TPRs when both of the classes are poisoned. 
When the attacking strength exceeds 4000, BIC-C-D fails to detect poisoned samples and falsely removes a large number of clean samples.
In short, compared with the other three defenses, our defense does not require any additional (clean) validation set or elaborate hyperparameter tuning, and it is more robust, especially when just one of the classes is corrupted. 

When there is no attack, our defense still sanitizes several components in both classes and falsely detects 789 clean training samples as poisoned samples. The average log likelihoods of these removed samples under ham and spam are $-879.77$ and $-852.88$, respectively (in other words, the likelihoods are similar).
Also, for an SVM trained on a ``perfectly sanitized'' dataset (\ie, clean dataset, without these samples), the classification accuracy on the set of these removed samples is only 0.5855. 
These falsely detected samples are close to the decision boundary and are well-explained by both classes. Given the similar likelihoods, it is BIC-efficacious to remove/revise the components formed by these samples; thus our defense inevitably produces some false positives. Besides, removing these ambiguous samples in fact slightly increases the test accuracy.

In Figure~\ref{fig:BIC-acc-vs-comp-spam}, we show how the total BIC cost, SVM test accuracy, and the number of detected poisoned samples change with the number of components removed/revised by our defense under the attack with 3000 poisoned samples injected into TREC05. (Note that in practice, we only remove the detected poisoned samples once the detection procedure terminates. But to see the algorithm's progress, via Figure~\ref{fig:BIC-acc-vs-comp-spam}, we trained an SVM on the training set without the detected samples at each detection step.) As we emphasized in Section \ref{method}, our method guarantees strict descent in the BIC objective. But we cannot guarantee that the test accuracy or the number of detected poisoned samples is non-decreasing.
% Different from component-wise decisions, sample-wise decisions may change along with the detection. 
Samples that were previously deemed poisoned might be restored clean at subsequent detection steps (re-assigned to the class to which they are labeled) and vice versa, which explains the slight fluctuation in the curves of SVM test accuracy and the number of detected poisoned samples. Overall, though, the strong trend of the two curves is an increase in detection accuracy and true positive detections with increasing detection steps. Specifically, the two curves increase sharply in the early stages and converge in the final stages, as the BIC cost is further decreased.

We show the number of components, number of revised components and number of removed components for both ham and spam under all attack cases in Table~\ref{tab:3}.
In general, the total number of removed and revised components increases as the attack is strengthened.
For most of the cases, our method prefers revising a poisoned component rather than removing it. A possible reason is that, given a large number of features (30000 for TREC05), it is difficult to cluster the clean samples and the poisoned samples into separate groups \cite{steinbach2004challenges}. 
Most of the poisoned components are formed by both clean and poisoned samples, and it is apparently most BIC-efficacious to revise them.

\section{Experiments on Multi-class Classification Tasks}\label{experiments_multi}

\subsection{Experiment Setup}\label{exp_setup_multi}
\textbf{Dataset and mixture model}: 
For multi-class ($W>2$) classification, we used the 20-Newsgroups dataset (20NG) \cite{20news}, MNIST \cite{MNIST}, CIFAR10 \cite{cifar10}, and STL10 \cite{STL10}. 
20NG collects news documents across 20 different news groups. Each group corresponds to a different topic and contains around 600 training samples and 400 test samples. MNIST is a dataset of 28x28 gray-scale images. It contains 5000 training images and 1000 test images per class. CIFAR10 consists of 32x32 color images, with 5000 training images and 1000 test images per class. STL10 is composed of 96x96 color images, with 500 training images and 800 test images\footnote{The STL10 dataset is an image recognition dataset with 100000 unlabeled images for developing unsupervised feature learning, deep learning, self-taught learning algorithms. Here we only use the labeled set.}. The experiments on each dataset involved 5 classes. For 20NG, we chose classes ``rec.sport.baseball'', ``soc.religion.christian'', ``comp.graphics'', ``rec.autos'', and ``misc.forsale''. For MNIST, CIFAR10, and STL10, we chose the first 5 classes. 

To simulate reasonable and potent embedded DP attacks, we used real samples as poisoning samples. For 20NG, we split the test set: 220 samples per class are used for testing and 160 samples per class are used for poisoning. For MNIST, we split the training set: 2000 images per class are used for training and 800 images per class are used for poisoning\footnote{For MNIST we only used half of the training samples.}. For CIFAR10, we split the training set: for each class, 4000 images are used for training and 800 images are preserved for poisoning. For STL10, we split the test set: for each class, 700 images are used for testing and 100 images are used for poisoning.  
For each dataset, the samples used for poisoning are injected into the training set, as described further below.

For each dataset, we trained a mixture model for each of its classes. We applied PMMs on 20NG. After pre-processing, the dictionary contains around 10000 unique words, and the training and test samples are represented using a bag-of-words. Each PMM component is a multinomial joint probability mass function. 
For MNIST, we applied Gaussian mixture models (GMMs). We first flattened the training images as 784-dimension vectors, normalized the intensity values to $[0,1]$, and centered the feature vectors. Then we trained GMMs on the pre-processed feature vectors. To reduce the model complexity, we assumed the features are independent conditioned on the mixture component of origin (\ie a diagonal covariance matrix for each Gaussian component).
For CIFAR10 and STL10, as the raw images are not very suitable for clustering, we trained GMMs on the feature vectors extracted from an internal layer of the victim NN-based classifier. In the experiments, we used the 512-dimensional penultimate layer features, which were again assumed independent, conditioned on the mixture component of origin. The features were again normalized and centered before GMM learning.

\textbf{DP attack and target classifiers}: 
For the multi-class classification task, we launched 5 DP attacks on all datasets. In attack $i=1,\dots,5$, we used samples of the first $i$ classes for poisoning. Samples from a given class $c$ are used to poison all other classes. That is,
% For attack $i=1,\dots,5$, we evenly distribute the 160 poisoning samples of class $c=1,\dots,i$ to class $k\neq c$ and label them as $k$.

For attack $i=1,\dots,5$:
\begin{itemize}[nosep]
    \item[] For class $c=1,\dots,i$:
    \begin{itemize}
        \item[] Evenly distribute the poisoning samples of class $c$ to the training sets of classes $w\neq c$ and label these samples as $w$.
    \end{itemize}
\end{itemize}

We chose linear SVM and LR as the target classifiers for 20NG and MNIST, as they are not effective for classifying complicated images such as CIFAR10 and STL10. For the NN based classifier, we chose a bi-directional one-layer LSTM recurrent neural network with 128 hidden units for 20NG, ResNet-18 \cite{ResNet} for MNIST and CIFAR10, and ResNet-34 for STL10\footnote{Since the number of labeled training samples of STL10 is not sufficient for training a complicated NN from scratch, we fine-tuned a pre-trained ResNet-34 on STL10.}.

\textbf{Evaluation criteria}: 
The performance of our defense (BIC-D) was assessed using the same metrics as in Section \ref{experiments_bi}: 1) improvement in test classification accuracy, relative to that of the poisoned classifier, of the classifier trained following data sanitization; 2) TPR; and 3) FPR.
We also applied the KNN-based defense (KNN-D) \cite{PaudiceML18}\footnote{Although the method was proposed for binary classification tasks, it is straightforward to apply it for multi-class classification tasks.}, GS-based defense (GS-D) \cite{Hong20GS}, and the SVD-based defense (SVD-D) \cite{DiakonikolasKK019}, using the same poisoned training sets. As \cite{mixture} was only proposed for binary classification (and assumes the poisoned class is known), we did not apply this method on the multi-class classification task. We also did not apply the SVD-based defense on the LSTM classifier, since it is only applicable to linear classifiers. For ResNet models, we applied the SVD-based defense on the output layer, since it is actually a linear classifier built on features extracted based on the previous layers.

\textbf{Hyper-parameter setting:} 
1) For the KNN-based defense, we set the number of neighbors at $K=10$, which is suggested by \cite{PaudiceML18}. 2) For the GS-based defense, we applied a DP-SGD optimizer with clip-norm of 2.0 and noise multiplier of 0.1 for training, which were suggested in \cite{Hong20GS}. 3) For the SVD-based defense, we set the number of detection steps at $\beta=2$, which is the same as in \cite{DiakonikolasKK019}.  To show the best performance for that method, we set $\epsilon$ to the real poisoning ratio if the training set is poisoned, and set it to 0.01 if there is no poisoning.

\subsection{Experimental Results}
The results are listed in Table~\ref{tab:20NG_acc}-\ref{tab:STL10_TPR}.
Table~\ref{tab:20NG_acc}, \ref{tab:MNIST_acc}, \ref{tab:CIFAR10_acc}, and \ref{tab:STL10_acc} show the test accuracy of victim classifiers as a function of attack strength on poisoned and sanitized 20NG, MNIST, CIFAR10, and STL10 datasets, respectively. 
We first trained the target classifiers on the attack-free datasets to get baseline test accuracies (\ie, column 0 of poisoned classifiers). 
Then we trained the classifiers on training sets poisoned by the 5 DP attacks described in Section \ref{exp_setup_multi}, with the resulting test accuracies in columns 1-5 of ``Poisoned", respectively.
For 20NG and MNIST, as the number of classes used for poisoning is increased to 5, the classification accuracies of SVM and LR drop by over 30\% (absolute percentage drop). The test accuracy of the NN-based classifier drops by nearly 30\% on 20NG and 20\% on MNIST. The test accuracies of ResNets drop by over 10\% on CIFAR10 and nearly 10\% on STL10.
Thus, the attacks designed in Section \ref{exp_setup_multi} are indeed pretty effective poisoning attacks against all target classifiers, on all datasets.

Then we applied our defense and KNN/GS/SVD-based defenses on the poisoned training sets and retrained the target classifiers on the sanitized datasets. The corresponding test accuracies on 20NG, MNIST, CIFAR10, and STL10 are shown as BIC-D, KNN-D, GS-D, and SVD-D in Table~\ref{tab:20NG_acc}, \ref{tab:MNIST_acc}, \ref{tab:CIFAR10_acc}, and \ref{tab:STL10_acc}, respectively. 
For 20NG, CIFAR10, and STL10, our defense outperforms the other three defenses in classification accuracy (marked in bold) in all attacking cases, excluding attack 0 (attack-free) against CIFAR10 and STL10. 
When there is no poisoning on CIFAR10 and STL10, SVD-D performs the best since we set $\epsilon$ as a small number. However, in practice, it will be difficult to find an appropriate value for $\epsilon$ in the absence of a clean validation set.
For 20NG, although KNN/GS/SVD-based defenses improve the test accuracies for most cases, the test accuracies drop by 10\%-25\% as the attack strength increases, compared with the clean baseline. Our BIC-based defense performs well and stably -- the test accuracy drops by 5\% at most.
For CIFAR10, the ResNet-18 with KNN/GS-based defense performs even worse than the poisoned classifier in all attacking cases. The SVD-based defense only improves the test accuracies by 4\% at most. By contrast, the test accuracy of the classifier with our method drops by only 2\% under the strongest attack, compared with the clean baseline.
For STL10, the ResNet-34 with KNN-based defense performs worse than the poisoned classifier in all attacking cases. The GS-based defense mitigates the negative effect of DP, but the test accuracy still drops by 6\% compared with the clean baseline. The SVD-based defense has little effect in improving test accuracy, compared with that of the poisoned classifier. On the other hand, the test accuracies of the classifier with our method drop by 1.31\% at most, compared with the clean baseline.

For MNIST, our method still performs well and beats GS/SVD-based defenses, while the KNN-based defense with $K=10$ (KNN-10-D in Table~\ref{tab:MNIST_acc}) gives slightly better results than ours. $K=10$ was suggested by \cite{PaudiceML18} and was chosen according to the performance of the algorithm evaluated on a (clean)
validation dataset of MNIST. However, for the embedded DP attack scenario considered here, the defender does not have access to a trusted validation set. Also, as we discussed in Section \ref{related work}, the choice of $K$ tremendously impacts the detector's performance. Thus, we also evaluated the performance of the KNN-based detector with $K=3$, with the resulting test accuracies shown as KNN-3-D in Table~\ref{tab:MNIST_acc}. As we can see, the test accuracy drops significantly under attack 5, compared with KNN-10-D and our method.
On the attack-free training set, the SVD-based defense performs best on SVM and LR, and the KNN-based defense performs best on ResNet-18, since the hyper-parameters of both methods are well-chosen. 

Table~\ref{tab:20NG_TPR}, \ref{tab:MNIST_TPR}, \ref{tab:CIFAR10_TPR}, and \ref{tab:STL10_TPR} show the TPRs and FPRs of the BIC/KNN/SVD-based defenses (since the GS-based defense does not identify the poisoned samples) on 20NG, MNIST, CIFAR10, and STL10 datasets, respectively. Since the performance of the SVD-based defense depends on the classifier architecture and training loss function (it evaluates the gradients of same), we respectively show its TPR/FPR on SVM, LR, and ResNet as SVD-D-S, SVD-D-L, and SVD-D-R.
We reiterate here that both KNN and SVD-based defenses evaluated here are supervised methods, \ie with appropriately chosen hyper-parameters.
For 20NG, CIFAR10, and STL10, compared with the other two defenses, our defense has relatively high TPRs and low FPRs for all cases. Almost no clean samples are falsely reported by our defense, while a large number of poisoned samples are correctly identified. The KNN-based defense falsely detects lots of clean samples in all attack cases, even when there is no poisoning. By contrast, the SVD-based defense only detects a small amount of poisoned samples, especially when the attack strength is weak.
For MNIST, the SVD-based detector has lower TPRs and FPRs than ours on SVM and LR, and does not perform well on ResNet.
The KNN-based detector with $K=10$ has higher TPRs and lower FPRs than our method under most attacking cases. 
Again, the performance of the KNN-based detector is affected by the choice of $K$, and $K=10$ was chosen based on the detector's performance evaluated on a clean validation set for MNIST. With $K=3$, the detector is less ``agressive'' -- it reports fewer detected poisoned images and has much lower TPRs.

Similar to the results in Section \ref{experiments_bi}, our method also falsely removes a few clean samples from the attack-free datasets (attack 0). These samples are well-explained by more than one class, and it is BIC-efficacious to re-distribute these samples. Removing these samples slightly {\it increases} the test accuracy, compared with the clean classifier baselines. In summary, our defense is an unsupervised method and significantly improves test accuracies for all classifiers on all datasets even under strong attacks, restoring the test accuracies close to the clean baselines.

\begin{table}[!t] 
\centering
% \scriptsize
% \footnotesize
% \small
\begin{tabular}{p{1.38cm}|p{0.75cm}p{0.75cm}p{0.75cm}p{0.75cm}p{0.75cm}p{0.75cm}}
\toprule
\hline
Attack & 0 & 1 & 2 & 3 & 4 & 5 \\
% \midrule
\hline
\multicolumn{7}{c}{\textbf{SVM}}\\
\hline
Poisoned & 0.9172 & 0.8681 & 0.7936 & 0.7036 & 0.6072 & 0.5281 \\
BIC-D & \textbf{0.9254} & \textbf{0.9127} & \textbf{0.9027} & \textbf{0.8845} & \textbf{0.8818} & \textbf{0.8736} \\
KNN-D & 0.9009 & 0.8891 & 0.8672 & 0.8236 & 0.7736 & 0.7391 \\
GS-D & 0.9127 & 0.8918 & 0.8691 & 0.8391 & 0.8073 & 0.7645 \\
SVD-D & 0.9181 & 0.8663 & 0.8272 & 0.7745 & 0.7254 & 0.6600 \\
\hline
\multicolumn{7}{c}{\textbf{LR}}\\
\hline
Poisoned & 0.9309 & 0.8781 & 0.8291 & 0.7536 & 0.6718 & 0.5909 \\
BIC-D & \textbf{0.9354} & \textbf{0.9218} & \textbf{0.9172} & \textbf{0.8954} & \textbf{0.8872} & \textbf{0.8809} \\
KNN-D & 0.8909 & 0.8791 & 0.8681 & 0.8309 & 0.7963 & 0.7700 \\
GS-D & 0.9181 & 0.8873 & 0.8855 & 0.8536 & 0.8373 & 0.7973 \\
SVD-D & 0.9309 & 0.9081 & 0.8690 & 0.8618 & 0.8463 & 0.8400 \\
\hline
\multicolumn{7}{c}{\textbf{LSTM}}\\
\hline
Poisoned & 0.8063 & 0.7427 & 0.7163 & 0.6736 & 0.6055 & 0.5336 \\
BIC-D & \textbf{0.8073} & \textbf{0.8064} & \textbf{0.8018} & \textbf{0.7800} & \textbf{0.7627} & \textbf{0.7481} \\
KNN-D & 0.7454 & 0.7409 & 0.7200 & 0.7136 & 0.7000 & 0.6664 \\
GS-D & 0.2636 & 0.2555 & 0.2400 & 0.2282 & 0.2545 & 0.2536 \\
% SVD-D & - & - & - & - & - & - \\
\hline
\bottomrule
\end{tabular}
\caption{Test set classification accuracy of victim classifiers on poisoned and sanitized 20NG datasets, under different attacks.}
\label{tab:20NG_acc}
% \vspace{-6mm}
\end{table}

\begin{table}[!t] 
\centering
% \scriptsize
% \footnotesize
\begin{tabular}{p{1.38cm}|p{0.75cm}p{0.75cm}p{0.75cm}p{0.75cm}p{0.75cm}p{0.75cm}}
\toprule
\hline
Attack & 0 & 1 & 2 & 3 & 4 & 5 \\
% \midrule
\hline
\multicolumn{7}{c}{\textbf{True Positive Rates (TPRs)}}\\
\hline
BIC-D & - & \textbf{0.8325} & \textbf{0.8203} & \textbf{0.7958} & \textbf{0.7843} & \textbf{0.7762} \\
KNN-D & - & 0.7687 & 0.7412 & 0.7296 & 0.7281 & 0.7104  \\
SVD-D-S & - & 0.2500 & 0.4156 & 0.4708 & 0.4828 & 0.4900 \\
SVD-D-L & - & 0.5937 & 0.6218 & 0.7020 & 0.7562 & 0.7650 \\
\hline
\multicolumn{7}{c}{\textbf{False Positive Rates (FPRs)}}\\
\hline
BIC-D & \textbf{0.0084} & \textbf{0.0145} & \textbf{0.0135} & \textbf{0.0145} & \textbf{0.0168} & \textbf{0.0155} \\
KNN-D & 0.1993 & 0.2001 & 0.1991 & 0.1997 & 0.2017 & 0.2091 \\
SVD-D-S & 0.01 & 0.0405 & 0.0632 & 0.0858 & 0.1118 & 0.1378 \\
SVD-D-L & 0.01 & 0.0219 & 0.0408 & 0.0483 & 0.0527 & 0.0581 \\
\hline
\bottomrule
\end{tabular}
\caption{TPRs and FPRs of two defenses on the 20NG dataset under all attack cases.}
\label{tab:20NG_TPR}
% \vspace{-6mm}
\end{table}

\begin{table}[!t] 
\centering
% \scriptsize
% \footnotesize
% \small
\begin{tabular}{p{1.38cm}|p{0.75cm}p{0.75cm}p{0.75cm}p{0.75cm}p{0.75cm}p{0.75cm}}
\toprule
\hline
Attack & 0 & 1 & 2 & 3 & 4 & 5 \\
% \midrule
\hline
\multicolumn{7}{c}{\textbf{SVM}}\\
\hline
Poisoned & 0.9621 & 0.8791 & 0.8717 & 0.8665 & 0.7773 & 0.5711 \\
BIC-D & 0.9536 & 0.9519 & 0.9551 & 0.9569 & 0.9537 & 0.9441 \\
KNN-10-D & 0.9560 & \textbf{0.9583} & \textbf{0.9564} & \textbf{0.9591} & \textbf{0.9618} & \textbf{0.9544} \\
KNN-3-D & 0.9616 & 0.9416 & 0.9367 & 0.9019 & 0.8772 & 0.8464 \\
GS-D & 0.9508 & 0.8846 & 0.8007 & 0.7867 & 0.7048 & 0.6213 \\
SVD-D & \textbf{0.9624} & 0.9537 & 0.9497 & 0.9415 & 0.9377 & 0.9239 \\
\hline
\multicolumn{7}{c}{\textbf{LR}}\\
\hline
Poisoned & 0.9606 & 0.8927 & 0.8521 & 0.8005 & 0.6592 & 0.6390 \\
BIC-D & 0.9628 & 0.9569 & 0.9508 & 0.9583 & 0.9552 & 0.9455 \\
KNN-10-D & 0.9636 & \textbf{0.9584} & \textbf{0.9560} & \textbf{0.9611} & \textbf{0.9565} & \textbf{0.9534} \\
KNN-3-D & 0.9604 & 0.9450 & 0.9359 & 0.8927 & 0.8814 & 0.8484 \\
GS-D & 0.9545 & 0.9241 & 0.8004 & 0.7186 & 0.6079 & 0.5754 \\
SVD-D & \textbf{0.9659} & 0.9536 & 0.9452 & 0.9377 & 0.9353 & 0.9392 \\
\hline
\multicolumn{7}{c}{\textbf{ResNet-18}}\\
\hline
Poisoned & 0.9976 & 0.9548 & 0.8986 & 0.8735 & 0.8597 & 0.8266 \\
BIC-D & 0.9986 & 0.9918 & 0.9951 & 0.9908 & 0.9911 & 0.9869 \\
KNN-10-D & \textbf{0.9988} & \textbf{0.9988} & \textbf{0.9976} & \textbf{0.9964} & \textbf{0.9961} & \textbf{0.9953} \\
KNN-3-D & 0.9974 & 0.9935 & 0.9706 & 0.9688 & 0.9622 & 0.9568 \\
GS-D & 0.9968 & 0.9787 & 0.9311 & 0.8846 & 0.8246 & 0.8001 \\
SVD-D & 0.9986 & 0.9920 & 0.9644 & 0.9322 & 0.9073 & 0.8433 \\
\hline
\bottomrule
\end{tabular}
\caption{Test set classification accuracy of victim classifiers on poisoned and sanitized MNIST datasets, under different attacks.}
\label{tab:MNIST_acc}
% \vspace{-6mm}
\end{table}

\begin{table}[!t] 
\centering
% \scriptsize
% \footnotesize
\begin{tabular}{p{1.38cm}|p{0.75cm}p{0.75cm}p{0.75cm}p{0.75cm}p{0.75cm}p{0.75cm}}
\toprule
\hline
Attack & 0 & 1 & 2 & 3 & 4 & 5 \\
% \midrule
\hline
\multicolumn{7}{c}{\textbf{True Positive Rates (TPRs)}}\\
\hline
BIC-D & - & 0.9562 & 0.9556 & 0.9429 & 0.9568 & 0.9315 \\
KNN-10-D & - & \textbf{0.9950} & \textbf{0.9918} & \textbf{0.9867} & \textbf{0.9834} & \textbf{0.9827} \\
KNN-3-D & - & 0.8662 & 0.8106 & 0.7833 & 0.7543 & 0.7397 \\
SVD-D-S & - & 0.8600 & 0.8418 & 0.8591 & 0.8821 & 0.8832 \\
SVD-D-L & - & 0.8812 & 0.8668 & 0.8875 & 0.9009 & 0.9082 \\
SVD-D-R & - & 0.7725 & 0.6550 & 0.5287 & 0.4581 & 0.4352 \\
\hline
\multicolumn{7}{c}{\textbf{False Positive Rates (FPRs)}}\\
\hline
BIC-D & 0.0465 & 0.0723 & 0.0503 & 0.0406 & 0.0561 & 0.0531\\
KNN-10-D & 0.0182 & 0.0175 & \textbf{0.0166} & \textbf{0.0169} & \textbf{0.0186} & \textbf{0.0194} \\
KNN-3-D & 0.0088 & 0.0109 & 0.0191 & 0.0394 & 0.0533 & 0.0695 \\
SVD-D-S & \textbf{0.0100} & 0.0112 & 0.0253 & 0.0338 & 0.0377 & 0.0467 \\
SVD-D-L & 0.0100 & \textbf{0.0095} & 0.0213 & 0.0270 & 0.0317 & 0.0367 \\
SVD-D-R & 0.0100 & 0.0182 & 0.0552 & 0.1131 & 0.1734 & 0.2259 \\
\hline
\bottomrule
\end{tabular}
\caption{TPRs and FPRs of two defenses on MNIST under all attack cases.}
\label{tab:MNIST_TPR}
\end{table}

\begin{table}[!t] 
\centering
% \scriptsize
% \footnotesize
% \small
\begin{tabular}{p{1.38cm}|p{0.75cm}p{0.75cm}p{0.75cm}p{0.75cm}p{0.75cm}p{0.75cm}}
\toprule
\hline
Attack & 0 & 1 & 2 & 3 & 4 & 5 \\
\hline
Poisoned & 0.8634 & 0.8502 & 0.8302 & 0.7974 & 0.7634 & 0.7430 \\
BIC-D & 0.8638 & \textbf{0.8616} & \textbf{0.8528} & \textbf{0.8452} & \textbf{0.8446} & \textbf{0.8416} \\
KNN-D & 0.7150 & 0.7154 & 0.6688 & 0.6758 & 0.6602 & 0.6752 \\
GS-D & 0.8272 & 0.8074 & 0.7866 & 0.7288 & 0.7036 & 0.6852 \\
SVD-D & \textbf{0.8668} & 0.8584 & 0.8466 & 0.8164 & 0.8046 & 0.7812 \\
\hline
\bottomrule
\end{tabular}
\caption{Test set classification accuracy of ResNet-18 on poisoned and sanitized CIFAR10 datasets, under different attacks.}
\label{tab:CIFAR10_acc}
% \vspace{-6mm}
\end{table}

\begin{table}[!t] 
\centering
% \scriptsize
% \footnotesize
\begin{tabular}{p{1.38cm}|p{0.75cm}p{0.75cm}p{0.75cm}p{0.75cm}p{0.75cm}p{0.75cm}}
\toprule
\hline
Attack & 0 & 1 & 2 & 3 & 4 & 5 \\
% \midrule
\hline
\multicolumn{7}{c}{\textbf{True Positive Rates (TPRs)}}\\
\hline
BIC-D & - & \textbf{0.9275} & \textbf{0.9263} & \textbf{0.9133} & \textbf{0.9378} & \textbf{0.9290} \\
KNN-D & - & 0.9025 & 0.8050 & 0.8112 & 0.7922 & 0.8010 \\
SVD-D & - & 0.3650 & 0.2662 & 0.3587 & 0.4171 & 0.3655 \\
\hline
\multicolumn{7}{c}{\textbf{False Positive Rates (FPRs)}}\\
\hline
BIC-D & 0.0267 & 0.0494 & 0.0717 & 0.0881 & 0.1405 & 0.1626 \\
KNN-D & 0.4596 & 0.4628 & 0.4514 & 0.4533 & 0.4545 & 0.4484 \\
SVD-D & \textbf{0.0100} & \textbf{0.0254} & \textbf{0.0587} & \textbf{0.0769} & \textbf{0.0932} & \textbf{0.1269} \\
\hline
\bottomrule
\end{tabular}
\caption{TPRs and FPRs of two defenses on CIFAR10 under all attack cases.}
\label{tab:CIFAR10_TPR}
% \vspace{-6mm}
\end{table}

\begin{table}[!t] 
\centering
% \scriptsize
% \footnotesize
% \small
\begin{tabular}{p{1.38cm}|p{0.75cm}p{0.75cm}p{0.75cm}p{0.75cm}p{0.75cm}p{0.75cm}}
\toprule
\hline
Attack & 0 & 1 & 2 & 3 & 4 & 5 \\
\hline
Poisoned & 0.9028 & 0.8902 & 0.8722 & 0.8548 & 0.8328 & 0.8165 \\
BIC-D & 0.9028 & \textbf{0.9080} & \textbf{0.9085} & \textbf{0.9020} & \textbf{0.9068} & \textbf{0.8897} \\
KNN-D & 0.8000 & 0.7825 & 0.7440 & 0.7391 & 0.7217 & 0.7065 \\
GS-D & 0.9008 & 0.8972 & 0.8882 & 0.8631 & 0.8508 & 0.8405 \\
SVD-D & \textbf{0.9068} & 0.8954 & 0.8714 & 0.8622 & 0.8282 & 0.8111 \\
\hline
\bottomrule
\end{tabular}
\caption{Test set classification accuracy of ResNet-34 on poisoned and sanitized STL10 datasets, under different attacks.}
\label{tab:STL10_acc}
% \vspace{-6mm}
\end{table}

\begin{table}[!t] 
\centering
% \scriptsize
% \footnotesize
\begin{tabular}{p{1.38cm}|p{0.75cm}p{0.75cm}p{0.75cm}p{0.75cm}p{0.75cm}p{0.75cm}}
\toprule
\hline
Attack & 0 & 1 & 2 & 3 & 4 & 5 \\
% \midrule
\hline
\multicolumn{7}{c}{\textbf{True Positive Rates (TPRs)}}\\
\hline
BIC-D & - & 0.9100 & \textbf{0.8850} & \textbf{0.8700} & \textbf{0.8575} & \textbf{0.8040} \\
KNN-D & - & \textbf{0.9700} & 0.8500 & 0.8133 & 0.8100 & 0.8040 \\
SVD-D & - & 0.3800 & 0.2800 & 0.3066 & 0.3050 & 0.2220 \\
\hline
\multicolumn{7}{c}{\textbf{False Positive Rates (FPRs)}}\\
\hline
BIC-D & \textbf{0} & \textbf{0.0012} & \textbf{0.0008} & \textbf{0.0008} & \textbf{0.0012} & \textbf{0.0028}  \\
KNN-D & 0.5028 & 0.4964 & 0.4936 & 0.4976 & 0.4944 & 0.4856 \\
SVD-D & 0.0100 & 0.0248 & 0.0576 & 0.0832 & 0.1112 & 0.1556 \\
\hline
\bottomrule
\end{tabular}
\caption{TPRs and FPRs of two defenses on STL10 dataset under all attack cases.}
\label{tab:STL10_TPR}
% \vspace{-6mm}
\end{table}

\begin{figure}
    \centering
    \begin{subfigure}[b]{0.4\textwidth}
        \centering
        \includegraphics[width=\textwidth]{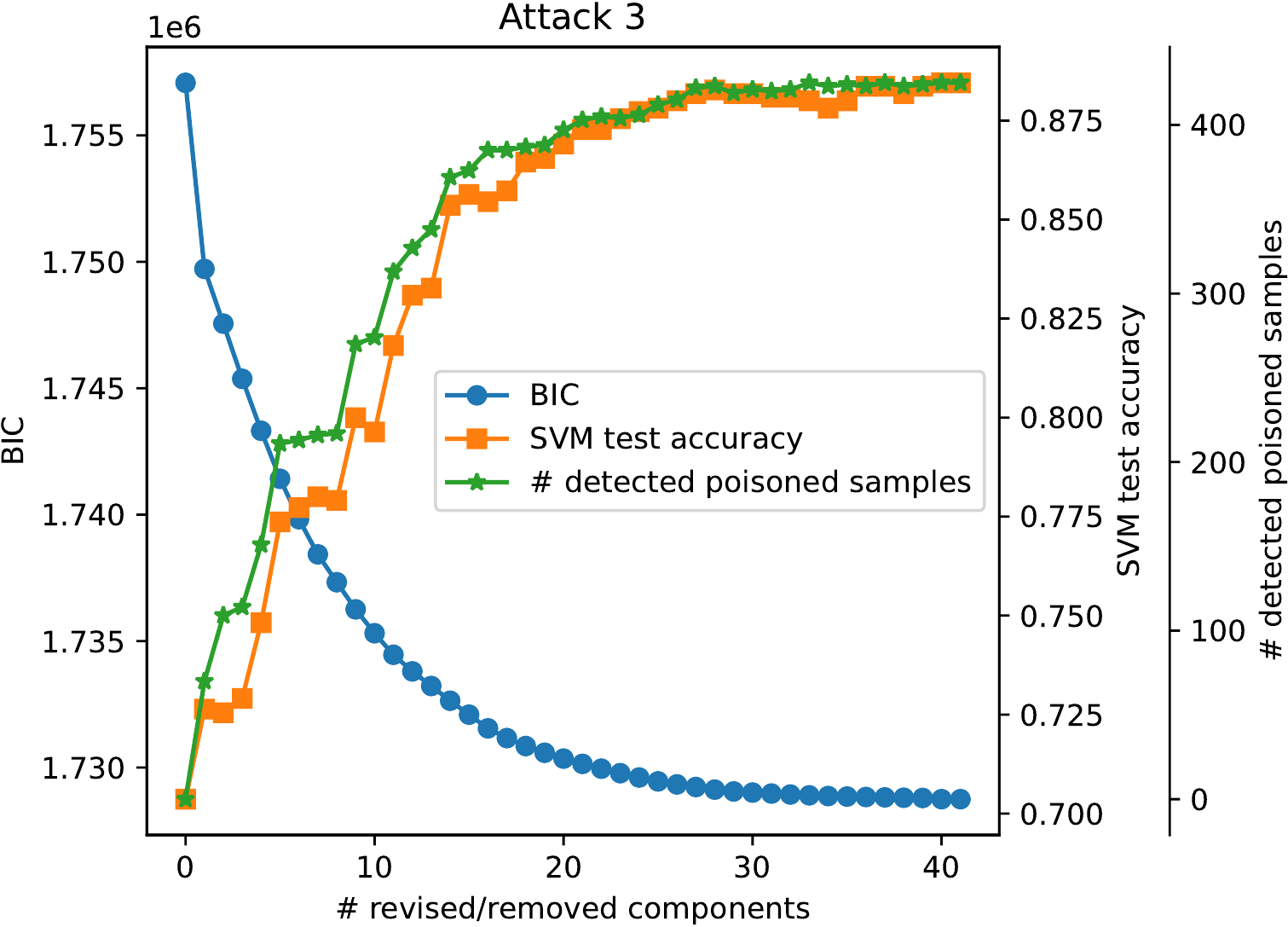}
        \caption{}
        \label{fig:20news-3}
    \end{subfigure}
    \begin{subfigure}[b]{0.4\textwidth}
        \centering
        \includegraphics[width=\textwidth]{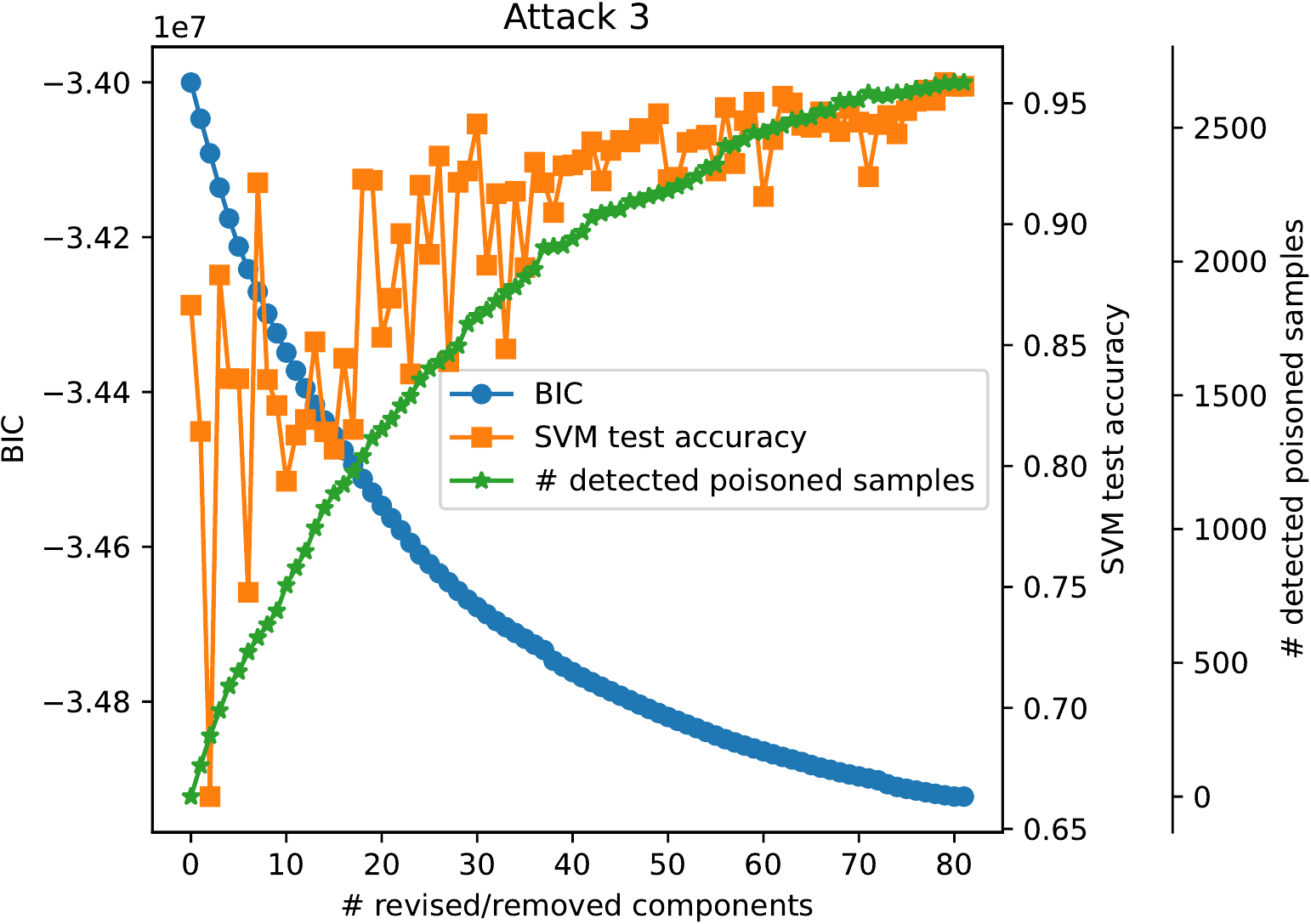}
        \caption{}
        \label{fig:MNIST-3}
    \end{subfigure}
    \caption{BIC cost, SVM test accuracy, and the number of detected poisoned samples as a function of the number of visited components (detection stages) under attack 3 against (a) 20NG dataset (b) MNIST dataset.}
    \label{fig:BIC-acc-vs-comp}
\end{figure}

In Figure~\ref{fig:20news-3} and \ref{fig:MNIST-3}, we respectively show how the total BIC cost, SVM test accuracy, and the number of detected poisoned samples change with the number of components removed/revised by our defense method under attack 3 against 20NG and MNIST.
(Note that in practice, we only remove the detected poisoned samples after the detection procedure terminates. Solely for visualization purposes, we trained an SVM on the training set without the detected samples at each detection step.) 
As we emphasized in Section \ref{method} and similar to the results in Section \ref{experiments_bi}, our method strictly descends in the BIC objective. But we cannot guarantee that the test accuracy or the number of detected poisoned samples is non-decreasing. 
For 20NG, both SVM test accuracy and the number of detected poisoned samples are almost strictly non-decreasing, with rapid increases during the first few detection steps.
For MNIST, the SVM test accuracy fluctuates heavily as the number of visited components increases and finally converges at around 0.95, while the number of detected poisoned samples is nearly monotonically increasing.

We show the number of components, number of revised components, and number of removed components of each class under all attack cases on all datasets in Table~\ref{tab:comp_multi}. 
For all datasets, the total number of removed and revised components is nearly strictly increasing as the attack is strengthened. For CIFAR10 and STL10, we applied our defense method on the features extracted from the penultimate layer of the NN-based classifier. 
The feature vectors of poisoned samples are well separated from those of clean samples; thus our method prefers removing poisoned components rather than revising them. By contrast, our detector prefers revising poisoned components rather than removing them for 20NG. Similar to the results for TREC05 in Section \ref{experiments_bi}, given a high feature dimensionality, most of the poisoned components are formed by both clean and poisoned samples, and it is apparently BIC-efficacious to revise them. 

\begin{table*}[!t] 
\centering
% \scriptsize
% \footnotesize
\begin{tabular}{l|llllll}
\toprule
\hline
Attack & 0 & 1 & 2 & 3 & 4 & 5 \\
% \midrule
\hline
\multicolumn{7}{c}{\textbf{20NG}}\\
\hline
\# components & (8,14,9,12,8) & (12,14,11,15,12) & (12,12,14,16,11) & (20,12,12,16,12) & (20,13,16,16,11) & (20,19,13,17,16) \\
\# revised components & (1,1,4,0,3) & (2,5,7,6,8) & (6,6,10,5,6) & (7,5,7,7,7) & (9,6,13,5,9) & (7,7,8,7,8) \\
\# removed components & (0,0,0,0,0) & (0,1,1,1,1) & (0,1,1,3,0) & (0,3,1,3,1) & (1,2,1,2,2) & (2,2,2,5,2) \\
\hline
\multicolumn{7}{c}{\textbf{MNIST}}\\
\hline
\# components & (28,28,27,27,28) & (28,31,30,24,33) & (44,31,39,37,43) & (38,41,39,38,44) & (31,33,33,38,47) & (33,40,41,35,42) \\
\# revised components & (6,4,15,10,5) & (4,2,12,15,7) & (3,4,14,14,5) & (3,7,7,3,4) & (6,4,12,11,4) & (7,5,2,7,6) \\
\# removed components & (0,0,2,1,0) & (0,7,5,3,4) & (5,8,10,8,11) & (6,17,9,11,13) & (8,16,9,12,18) & (9,19,15,12,18) \\
\hline
\multicolumn{7}{c}{\textbf{CIFAR10}}\\
\hline
\# components & (13,14,12,15,11) & (14,15,21,17,20) & (15,13,19,19,23) & (19,13,18,17,19) & (21,16,17,14,22) & (14,19,16,16,17) \\
\# revised components & (1,2,4,3,1) & (0,0,1,0,1) & (0,0,0,1,1) & (0,0,0,0,1) & (0,0,0,0,0) & (0,0,0,0,0) \\
\# removed components & (0,0,1,1,0) & (3,2,3,2,2) & (5,2,5,4,4) & (6,3,8,5,4) & (7,5,6,6,5) & (7,6,6,8,6) \\
\hline
\multicolumn{7}{c}{\textbf{STL10}}\\
\hline
\# components & (8,10,11,8,10) & (9,9,16,9,12) & (10,8,16,12,12) & (15,13,17,17,11) & (15,15,16,16,11) & (17,14,12,14,16) \\
\# revised components & (0,0,0,0,0) & (0,1,0,0,2) & (1,0,1,1,0) & (1,0,0,1,4) & (2,1,1,1,0) & (1,0,1,0,1) \\
\# removed components & (0,0,0,0,0) & (0,1,2,2,1) & (2,1,4,3,3) & (3,3,3,5,3) & (4,5,4,6,4) & (4,5,4,5,6) \\
\hline
\bottomrule
\end{tabular}
\caption{The number of components, number of revised components, and number of removed components of each class under all attack cases on 20NG, MINIST, CIFAR10, and STL10 datasets.}
\label{tab:comp_multi}
% \vspace{-6mm}
\end{table*}

\section{Computational Complexity of the BIC-based Defense}

For attack 1 poisoning the MNIST dataset, we report the computation time required to 
implement different defenses and that required to train the DNN
classifier (deep learning).
All the experiments are conducted on a compute platform consisting of Intel i9-10900K CPU, NVIDIA GeForce 3080 GPU and 32GB memory.
Table~\ref{tab:execution_time} shows the time used for training a ResNet-18 DNN classifier and deploying BIC/KNN/GS/SVD-based defenses.
We train the ResNet-18 for 50 epochs. The initial learning rate of 0.001 is  divided by 10 every 20 epochs.
For the BIC/KNN/SVD-based methods, the computational complexity is the time spent on anomaly detection and removal.
Since the GS-based defense mitigates DP attacks during DNN training, its reported computational time is the training time.
As we can see, KNN-based defense is extremely fast -- it detects and removes anomalies in about 2 seconds. Our BIC-based defense only sanitizes one component in one class at each detection step, and thus takes longer than the KNN-based defense. However, the execution time of our BIC-based defense is comparable to the DNN training time. As expected, SVD-based defense is computationally expensive as it performs SVD and re-trains the DNN at each detection step. GS-based defense greatly increases the training time due to the DP-SGD optimizer applied during DNN training.

\begin{table}[!t] 
\centering
% \scriptsize
% \footnotesize
% \small
\begin{tabular}{ccccc}
\toprule
\hline
ResNet-18 training & BIC-D & KNN-D & GS-D & SVD-D-R \\
\hline
497.67 & 427.32 & 2.08 & 1808.68 & 1122.42 \\
\hline
\bottomrule
\end{tabular}
\caption{Time (in seconds) used for training a ResNet-18 and deploying BIC/KNN/GS/SVD-based defenses on the MNIST dataset poisoned by attack 1.}
\label{tab:execution_time}
% \vspace{-6mm}
\end{table}

\section{Conclusion}\label{summary}

We proposed an \textit{unsupervised} BIC-based mixture model defense against DP attacks on classifiers, where poisoned samples are an unknown subset (potentially) simultaneously embedded in the training sets of multiple classes.  Our defense applies mixture modeling to accurately explain the poisoned dataset and concentrate poisoned samples into several mixture components. Also, it jointly identifies poisoned components and poisoned samples within them by minimizing the BIC cost, with the identified poisoned samples purged from the training set prior to classifier training. We launched our defense and three other defenses against DP attacks targeting SVM, LR, and NN-based classifiers for the TREC05, 20NG, MNIST, CIFAR10, and STL10 domains. Experiments demonstrate the effectiveness and robustness of our defense under strong attacks, as well as the superiority over the other defenses. 

\section*{Acknowledgements}
This research was supported in part by grants
from  AFOSR and ONR and by a gift from Cisco.

\bibliographystyle{plain}  
\bibliography{refs-full}
\end{document}